\documentclass{article}

\usepackage{PRIMEarxiv}

\usepackage[utf8]{inputenc} 
\usepackage[T1]{fontenc}    
\usepackage{hyperref}       
\usepackage{url}            
\usepackage{booktabs}       
\usepackage{amsfonts}       
\usepackage{nicefrac}       
\usepackage{microtype}      
\usepackage{lipsum}
\usepackage{fancyhdr}       
\usepackage{graphicx}       
\usepackage{subcaption}
\usepackage{amssymb, latexsym, amsmath, mathtools, amsthm, mathrsfs} 
\graphicspath{{media/}}     

\title{Applying Regularized Schrödinger-Bridge-Based Stochastic Process in Generative Modeling}

\author{
  Ki-Ung Song \\
  Department of Mathematical Sciences \\
  Seoul National University \\
  Seoul, Korea \\
  \texttt{sk851@snu.ac.kr} \\
}

\begin{document}
\maketitle

\begin{abstract}
Compared to the existing function-based models in deep generative modeling, the recently proposed diffusion models have achieved outstanding performance with a stochastic-process-based approach. But a long sampling time is required for this approach due to many timesteps for discretization. 
Schrödinger bridge (SB)-based models attempt to tackle this problem by training bidirectional stochastic processes between distributions. However, they still have a slow sampling speed compared to generative models such as generative adversarial networks. And due to the training of the bidirectional stochastic processes, they require a relatively long training time.
Therefore, this study tried to reduce the number of timesteps and training time required and proposed regularization terms to the existing SB models to make the bidirectional stochastic processes consistent and stable with a reduced number of timesteps. Each regularization term was integrated into a single term to enable more efficient training in computation time and memory usage. Applying this regularized stochastic process to various generation tasks, the desired translations between different distributions were obtained, and accordingly, the possibility of generative modeling based on a stochastic process with faster sampling speed could be confirmed. The code is available at \url{https://github.com/KiUngSong/RSB}.
\end{abstract}


\let\thefootnote\relax\footnotetext{This work was done as a dissertation submitted for the degree of Master of Science.}

\section{Introduction}
\label{sec:introduction}
As deep neural networks become essential elements in modern artificial intelligence research, various deep generative models and related neural network architectures have been proposed.
One of the most widely known deep generative models is generative adversarial networks (GANs) \cite{goodfellow2014generative}. They are based on adversarial training of generator and discriminator networks and have shown outstanding performance in various fields. 
Based on a log-likelihood of desired data distribution $\mathcal{P}$, variational autoencoders (VAEs) \cite{kingma2019introduction} and normalizing flows \cite{kingma2018glow} were proposed. VAEs are trained with a lower bound of the log-likelihood designed with encoder and decoder networks. And normalizing flows are trained with an invertible design of neural network architectures for the exact computation of the log-likelihood. Although there are differences in specific ways, they all approach generative modeling as a function.

Recently, diffusion models \cite{ho2020denoising, song2020score} have been proposed and shown outstanding performance with a stochastic-process-based approach.
Since the latent space of generative modeling, $\mathcal{Z}$, is assumed to be a Gaussian noise space, diffusion models first consider the stochastic noising process, say forward process, from $\mathcal{P}$ to $\mathcal{Z}$. Then, they consider a generation process as a corresponding backward stochastic process of the forward process.
Due to their impressive performance and formulation, they are applied in various fields, including largely pre-trained multimodal models \cite{ramesh2022hierarchical, saharia2022photorealistic}.

Under the neural network's universal approximation property, deep generative models have achieved remarkable improvement in various generation tasks such as unconditional generation, image-to-image translation, image super-resolution, \emph{etc.}
Even though the desired generation outcome varies for each situation, every situation is to find a proper mapping between two different desired distributions $\mathcal{P}$ and $\mathcal{Q}$ with task-dependent conditions.
For instance, in the case of an unconditional data generation task, $\mathcal{P}$ is the desired data distribution, and $\mathcal{Q}$ is the distribution of the latent space $\mathcal{Z}$, \emph{e.g.} Gaussian distribution. 
And, in the case of an image-to-image translation task, two different image domains $\mathcal{P}$ and $\mathcal{Q}$ are given, for instance, male and female. Then the main objective is to find the proper mapping between $\mathcal{P}$ and $\mathcal{Q}$ while preserving the semantic information of the given image, \emph{e.g.} identity of the human face.
There are various studies on generative model frameworks and related neural network architecture for each generation task. In other words, the existing methods on deep generative models rely heavily on what the two specific distributions $\mathcal{P}$ and $\mathcal{Q}$ are.

And among many generative models, the two main approaches are competing for the best performance: GANs and diffusion models. For years, GANs have shown an ability to generate high-quality images, and diffusion models demonstrated that they can be better than GANs \cite{dhariwal2021diffusion} in an image generation task.
However, each model has its shortcomings. In the case of GANs, various models suffer from unstable training and failure modes. Among the failure modes, there is a mode collapse problem where the trained models do not fully cover the desired data space. 
On the other hand, diffusion models show relatively stable training and high mode coverage performance. But the main disadvantage is its slow inference speed since it needs multiple timesteps to discretize the stochastic process.
But, the success of diffusion models provides a new idea in generative modeling. Since they consider the generation process as a series of stochastic processes rather than a single function, it demonstrated that the application of stochastic processes in generative modeling could achieve both high mode coverage performance and high-quality generation.

As mentioned above, various generation tasks depend on what the desired distributions $\mathcal{P}$ and $\mathcal{Q}$ are. And in many cases, there is no need for $\mathcal{Q} = \mathcal{Z}$. Even in generation situations such as text-conditional image generation, text-to-speech translation, or image captioning, the transformation between two distributions with different modalities should be considered.
Although diffusion models proposed a multi-stage stochastic-process-based generative modeling rather than a single-stage function, they depend on the forward process from $\mathcal{P}$ to $\mathcal{Z}$. Thus, they cannot construct bidirectional stochastic processes between arbitrary distributions $\mathcal{P}$ and $\mathcal{Q}$.
To tackle this problem, various conditioning methods \cite{choi2021ilvr, meng2021sdedit} have been proposed. But, these diffusion-based approaches still suffer from the slow sampling speed.
However, in a generative framework where stochastic processes between arbitrary distributions are constructed, it can be quite possible to improve the diffusion model's disadvantages while maintaining the advantage.

From the perspective of applied mathematics, the problem of transportation between two distributions $\mathcal{P}$ and $\mathcal{Q}$ with minimal cost can be expressed as an optimal transport (OT) problem.
And based on a Schrödinger bridge (SB) problem, which is an extension of entropy regularized OT problem, the desired bidirectional stochastic processes can be obtained.
Therefore, based on the SB problem's formulation, some generative modelings \cite{chen2022likelihood, de2021diffusion, wang2021deep} were proposed. And the recent work \cite{chen2022likelihood} has proposed an SB-based stochastic process as an extension to the diffusion model's stochastic process.
SB models require a relatively small number of timesteps compared to the diffusion models because the bidirectional processes are learnable. However, they still need a large number of evaluation steps than function-based generative modelings such as GANs.
Therefore, by modifying the existing SB-based formulation, this study tried to construct bidirectional stochastic processes with a reduced number of timesteps compared to the previous SB-based works.

\section{Backgrounds}
\label{sec:backgrounds}
Before the main discussion, backgrounds related to this work are presented in this chapter. First, basic concepts of OT and SB problem are briefly introduced with their formulation. The presented definitions and flow of explanation mainly referred to the work of Peyré \cite{peyre2017computational} and Vargas \cite{vargas2021machine}. Next, detailed backgrounds related to diffusion models are explained. And lastly, formulations of SB-based generative modeling are presented.

\subsection{Preliminaries}
Given two data spaces $X$ and $Y$, let $\mathcal{M}(X)$ and $\mathcal{M}(Y)$ be the set of probability measures, respectively. The optimal transport (OT) problem aims to formulate minimal-cost transportation from one data space to another. Let $T:X \rightarrow Y$ be a continuous map, then a corresponding push-forward operator $T_{\#}:\mathcal{M}(X) \rightarrow \mathcal{M}(Y)$ exists. Then Kantorovich's OT problem with continuous measures can be expressed as
\begin{equation}\label{eq:kantorovich}
\begin{gathered}
    \inf_{\pi \in U(\alpha, \beta)} \int_{X \times Y} c(x,y) d\pi(x,y), \\ 
    s.t. \quad U(\alpha, \beta) = \{ \pi \in \mathcal{M}(X \times Y) : \text{proj}_{X\#}\pi = \alpha \ and \ \text{proj}_{Y\#}\pi = \beta \},
\end{gathered}
\end{equation}

\noindent where $\text{proj}_{X\#}$ and $\text{proj}_{Y\#}$ is the push-forwards of the projections $\text{proj}_X(x,y)=x$ and $\text{proj}_Y(x,y)=y$ respectively and $\alpha$ and $\beta$ are probability measures on data space $X$ and $Y$ respectively..

By adding entropy regularization term in \ref{eq:kantorovich}, stochastic nature can implicitly conditioned to the OT problem. And it is given as below
\begin{equation}\label{eq:entropy_reg}
    \inf_{\pi \in U(\alpha, \beta)} \int_{X \times Y} c(x,y) d\pi(x,y) + \epsilon D_{KL}(\pi \vert \alpha \times \beta),
\end{equation}

\noindent where $D_{KL}(p \vert q) = \int_{X \times Y}\log{(\frac{dp}{dq})}dp$ is Kullback–Leibler (KL) divergence for distributions $p$ and $q$.

\noindent By refactoring the \ref{eq:entropy_reg} with Gibbs distribution $\mathcal{K}$ which is given as
\begin{equation}\label{eq:Gibbs}
d\mathcal{K}(x,y) = \exp{-\frac{c(x,y)}{\epsilon}}d\alpha(x)d\beta(y),
\end{equation}

\noindent the entropy regularized OT problem \ref{eq:entropy_reg} can be expressed as 
\begin{equation}\label{eq:SP}
    \inf_{\pi \in U(\alpha, \beta)} D_{KL}(\pi \vert \mathcal{K}).
\end{equation}

\noindent The above form of the problem is often called a static Schrödinger problem. This is a situation where there the Gibbs distribution $\mathcal{K}$ contains information about the cost function, and path $\pi$ between $\alpha$ and $\beta$ is optimized to be close with the Gibbs distribution as a reference.

By extending this, Schrödinger bridge (SB) problem can be proposed as
\begin{equation}\label{eq:SB}
    \inf_{\pi \in \mathcal{D}(\mathcal{P}, \mathcal{Q})} D_{KL}(\pi \vert \mathcal{W}),
\end{equation}

\noindent where reference measure $\mathcal{W}$ replaces the Gibbs measure and $\mathcal{D}(\mathcal{P}, \mathcal{Q})$ is a set of path measures with marginals of desired distribution $\mathcal{P}$ and $\mathcal{Q}$.
This formulates a more general situation of finding a path measure between $\mathcal{P}$ and $\mathcal{Q}$ where cost information is implicitly reflected in a choice of reference measure.
From the perspective of KL divergence as a distance, it can be interpreted that the process of reducing the distance between the path measure and reference measure reflects the nature of OT since the reference measure contains cost information.

The choice of $\mathcal{W}$ as a prior knowledge enables different interpretations of the SB problem. For instance, if $\mathcal{W}$ is uniform distribution, then \ref{eq:SB} becomes equivalent to \ref{eq:SP} with entropy. And it was demonstrated that the SB problem is equivalent to a stochastic control problem with a proper choice of $\mathcal{W}$ \cite{vargas2021machine, Pavon1991}. 
Let $\mathcal{W}^{\gamma}$ be the Wiener measure with volatility $\gamma$, then path measure $\pi \in \mathcal{D}(\mathcal{P}, \mathcal{Q})$ can be expressed as a distribution which evolves according to the solution of stochastic differential equations: forward direction and backward direction of Ito process form as
\begin{equation}\label{eq:sde}
\begin{gathered}
    dx_t = f_{t}dt + \sqrt{\gamma} dw_t, \\
    dx_t = b_{t}dt + \sqrt{\gamma} dw_t.
\end{gathered}
\end{equation}

\noindent With the above forward and backward Ito process, the SB problem can be expressed as the following two alternate objectives
\begin{equation}\label{eq:SB_obj1}
\begin{gathered}
    \min_{\pi \in \mathcal{D}(\mathcal{P}, \mathcal{Q})} D_{KL}(\pi \vert \mathcal{W}^{\gamma}) = \min_{f_t} \mathbf{E}_{\pi} \left[\int_{0}^{1} \frac{1}{2\gamma} \Vert f_t \Vert^2 dt \right], \\
    s.t. \quad dx_t = f_{t}dt + \sqrt{\gamma} dw_t,\ x_0 \sim \mathcal{P},\ x_1 \sim \mathcal{Q},
\end{gathered}
\end{equation}

\begin{equation}\label{eq:SB_obj2}
\begin{gathered}
    \min_{\pi \in \mathcal{D}(\mathcal{P}, \mathcal{Q})} D_{KL}(\pi \vert \mathcal{W}^{\gamma}) = \min_{b_t} \mathbf{E}_{\pi} \left[\int_{0}^{1} \frac{1}{2\gamma} \Vert b_t \Vert^2 dt \right], \\
    s.t. \quad dx_t = b_{t}dt + \sqrt{\gamma} dw_t,\ x_1 \sim \mathcal{Q},\ x_0 \sim \mathcal{P}
\end{gathered}
\end{equation}

\noindent The above objectives do not provide information about an update rule of drift $f_t$ and $b_t$ in a stochastic process. But it means that the SB problem can be formulated as an optimal control problem with bidirectional stochastic processes minimizing their energy.

\subsection{Diffusion Models}
Except for diffusion models, the previously mentioned generative models aim to train a one-stage function from $\mathcal{Z}$ to $\mathcal{P}$. In GANs, a mapping $G:\mathcal{Z} \rightarrow \mathcal{P}$ is trained directly. And in VAEs, two mappings are trained: encoder $E:\mathcal{P} \rightarrow \mathcal{Z}$ and decoder $D:\mathcal{Z} \rightarrow \mathcal{P}$. In normalizing flows, invertible network $G:\mathcal{P} \rightarrow \mathcal{Z}$ is trained where inference is done by $G^{-1}:\mathcal{Z} \rightarrow \mathcal{P}$.
However, diffusion models first considered a transition from $\mathcal{P}$ to $\mathcal{Z}$ as a stochastic forward process. The forward process is a noising process and it can be formulated in various ways. One can be formulated via the Markov process \cite{ho2020denoising} as
\begin{equation}\label{eq:ddpm_f}
    x_{t+1} = \mathcal{N}(\sqrt{\beta_t}x_t, (1-\beta_t)\mathbf{I}).
\end{equation}

Meanwhile, the forward noising process can be formulated as a continuous stochastic differential equation (SDE), unlike the above discrete Markov process. This was proposed in the score-based generative model (SGM) \cite{song2020score} which became the prototype of the diffusion model and its SDE form is expressed as
\begin{equation}\label{eq:sgm_f}
    dx_t = f_{t}(x_t)dt + g_tdw_t.
\end{equation}

\noindent With such a forward process is given, it is known that the corresponding backward stochastic process is expressed as
\begin{equation}\label{eq:sgm_b}
    dx_t = \left[ f_{t}(x_t) - g_t^2\nabla_x \log{p_t(x_t)} \right]dt + g_tdw_t,
\end{equation}

\noindent where $p_t$ is the marginal distribution of $x_t$, and $\nabla_x \log{p_t(x_t)}$ is called score function of $p_t$.
Since the numerical computation of the above SDEs requires discretization, with an Euler-Maruyama scheme and discrete-time step size $\gamma$, the numerical sampling process of the above two processes \ref{eq:sgm_f} and \ref{eq:sgm_b} is given as
\begin{equation}\label{eq:sgm_f_euler}
    x_{t+1} = x_{t} + \gamma f_{t}(x_t) + \sqrt{\gamma} g_t z,
\end{equation}
\begin{equation}\label{eq:sgm_b_euler}
    x_{t} = x_{t+1} - \gamma \left[f_{t+1}(x_{t+1}) - g_{t+1}^2\nabla_x \log{p_{t+1}(x_{t+1}}) \right] + \sqrt{\gamma} g_{t+1}z,
\end{equation}

\noindent where $z$ is Gaussian noise. The drift terms $f_t$ and $g_t$ are derived from the discrete-time diffusion frameworks such as DDPM \cite{ho2020denoising}. The previously mentioned discrete noising process \ref{eq:ddpm_f} of DDPM \cite{ho2020denoising} can be induced in a SDE form \ref{eq:sgm_f}. It is known as VPSDE and given as
\begin{equation}\label{eq:VPSDE}
    dx_t = - \frac{1}{2}\beta_t x_t dt + \sqrt{\beta_t}dw_t.
\end{equation}

\noindent Since the forward process \ref{eq:sgm_f} of SGM has a fixed linear drift, it is possible to compute an exact transition kernel that transits $x_0$ to $x_t$. The corresponding transition kernel of VPSDE is
\begin{equation}\label{eq:VPSDE_transtion}
    p(x_t|x_0) = \mathcal{N}\left( x_t ; x_0e^{- \frac{1}{2} \int_0^t\beta_s ds}, \left(1 - e^{-\int_0^t\beta_s ds} \right)\mathbf{I} \right).
\end{equation}

\noindent Considering the numerical backward process \ref{eq:sgm_b_euler} for generation, in addition to the information of drift $f_t$ and $g_t$, the information about score function $\nabla_x \log{p_t(x_t)}$ is also essential. Thus, the training of diffusion models is designed to make neural network $s_\theta(x_t,t)$ approximate the non-linear drift $\nabla_x \log{p_{t}(x_t})$ at each $x_t$. To achieve this, objective function for training is given as
\begin{equation}\label{eq:sgm_obj}
    \mathbf{E}_{x_{0}, t}\left[ \Vert s_\theta(x_t,t) - \nabla_{x} \log{p(x_t|x_0)} \Vert^2_2 \right].
\end{equation}

\noindent By minimizing the above objective, known as score-matching framework \cite{vincent2011connection}, the neural network $s_\theta(x_t,t)$ can approximate the desired score function $\nabla_x \log{p_{t}(x_t})$ properly.

Although diffusion models have shown remarkable performance and scalability to various domains, they suffer from slow sampling speed. Thus, many studies \cite{jolicoeur2021gotta, xiao2021tackling, vahdat2021score, kim2022maximum} have proposed methods to improve sampling speed.
These methods 1) utilize GAN's approximation ability, 2) introduce a faster numerical algorithm for solving SDEs, or 3) consider the diffusion process in the latent space, which is equivalent to diffusion models with non-linear drift $f_t$ and $g_t$ in the original data space.
However, even in these attempts, the diffusion models still require a large number of timesteps for inference compared to one-stage generative models such as GANs. Also, bidirectional stochastic processes between any desired distribution $\mathcal{P}$ and $\mathcal{Q}$ cannot be obtained in the framework of diffusion models.

As discussed before, a modality-agnostic transformation between $\mathcal{P}$ and $\mathcal{Q}$ is required to solve various real-world generation problems. For instance, at present, text-conditional image generation and image captioning are approached as separate tasks. But if the transformation between image and text data spaces are induced together as bidirectional processes, the two tasks can be learned as one framework.
Therefore, there is a need for stochastic-process-based generative modeling between any $\mathcal{P}$ and $\mathcal{Q}$. More research is still needed for the general modality-agnostic transformation between data distributions with different dimensions. 
But, the SB formulation can construct desired stochastic processes between data distributions of the same dimension, and it can be seen as an attempt at a more general generative framework while showing a direction for solving the limitations of the diffusion model.

\subsection{Schrödinger Bridge in Generative Modeling}
From the SB problem \ref{eq:SB} perspective, the work of SB-FBSDE \cite{chen2022likelihood} consists of two learnable generative processes: forward and backward stochastic processes. Similar to the  SGM's forward and backward process, \ref{eq:sgm_f} and \ref{eq:sgm_b}, SB-based formulation from the work of \cite{chen2022likelihood} is given as
\begin{equation}\label{eq:fbsde_f}
    dx_t = \left[ f_{t}(x_t) + g_t^2\nabla_x \log{\psi_t(x_t)} \right]dt + g_tdw_t,
\end{equation}
\begin{equation}\label{eq:fbsde_b}
    dx_t = \left[ f_{t}(x_t) - g_t^2\nabla_x \log{\hat{\psi}_t(x_t)} \right]dt + g_tdw_t,
\end{equation}

\noindent where $\nabla_x \log{\psi_t(x_t)}$ and $\nabla_x \log{\hat{\psi}_t(x_t)}$ are non-linear drift terms. If $\nabla_x \log{\psi_t(x_t)}=0$ holds, it can be readily confirmed that it is equivalent to that of \ref{eq:sgm_f} and \ref{eq:sgm_b}. With discretization step size $\gamma$ and Euler-Maruyama scheme, the sampling process for numerical computation is given as
\begin{equation}\label{eq:fbsde_f_euler}
    x_{t+1} = x_{t} + \gamma \left[f_{t}(x_t) + g_t^2\nabla_x \log{\psi_t(x_t)} \right] + \sqrt{\gamma} g_t z,
\end{equation}
\begin{equation}\label{eq:fbsde_b_euler}
    x_{t} = x_{t+1} - \gamma \left[f_{t+1}(x_{t+1}) - g_{t+1}^2\nabla_x \log{\hat{\psi}_{t+1}(x_{t+1})} \right] + \sqrt{\gamma} g_{t+1} z.
\end{equation}

\noindent The similarity between the forward and backward processes of SB-FBSDE and those of SGM is straightforward. And since SB-FBSDE has non-linear drift in both bidirectional processes unlike SGM, SB-based stochastic processes are a generalization of that of SGM.
While the drift term $f_t$ and $g_t$ are derived from discrete-time diffusion models in SGM's formulation, they are not induced theoretically in the SB-FBSDE formulation. Since there is no information to construct $f_t$ and $g_t$ in SB-FBSDE, they were manually set as those of VESDE or simply as $f_t=0$ and $g_t=1$.
And unlike diffusion models, the above SB formulation cannot be trained in the form of score-matching, so two processes \ref{eq:fbsde_f} and \ref{eq:fbsde_b} are transformed into an equivalent SDE problem to construct a loss objective. In this process, non-linear drift terms of SB-FBSDE have the following relationship with score function $\nabla_x \log{p^{\text{SB}}_{t}(x_t)}$ of path measure $p^{\text{SB}}_{t}$ of SB problem:
\begin{equation}\label{eq:fbsde_prop}
    \nabla_x \log{\psi_t(x_{t})} + \nabla_x \log{\hat{\psi}_t(x_{t})} = \nabla_x \log{p^{\text{SB}}_{t}(x_t)}.
\end{equation}

There is another SB-based formulation, Diffusion Schrödinger Bridge (DSB) \cite{de2021diffusion}, which was induced as Markov processes. It is defined as
\begin{equation}\label{eq:dsb_f}
\begin{gathered}
    p^f(x_{t+1} | x_{t}) = \mathcal{N}(x_{t+1}; x_{t} + \gamma \mathbf{f}_{t}(x_t), 2\gamma \mathbf{I})
    = \mathcal{N}(x_{t+1} ; F_{t}(x_{t}), 2\gamma \mathbf{I}),
\end{gathered}
\end{equation}
\begin{equation}\label{eq:dsb_b}
\begin{gathered}
    p^b(x_{t} | x_{t+1}) = \mathcal{N}(x_{t}; x_{t+1} + \gamma \mathbf{b}_{t+1}(x_{t+1}), 2\gamma \mathbf{I})
    = \mathcal{N}(x_t ; B_{t+1}(x_{t+1}), 2\gamma \mathbf{I}),
\end{gathered}
\end{equation}

\noindent for $t \in \{1,\dots,T-1 \}$ and $T$ is the number of total time step. By letting the approximation of $p^f(x_{t} | x_{t+1})$ as
\begin{equation}\label{eq:dsb_induce}
    p^f(x_{t} | x_{t+1}) \approx \mathcal{N}(x_{t}; x_{t+1} - \gamma \mathbf{f}_{t}(x_{t+1}) + 2\gamma \nabla_x \log{p^{f}_{t+1}(x_{t+1}}), 2\gamma \mathbf{I}),
\end{equation}

\noindent if the two processes $p^f(x_{t} | x_{t+1})$ and $p^b(x_{t} | x_{t+1})$ corresponds, the drift term of $p^b$ can be expressed with the drift term and score function of $p^f_t$. Thus, it leads to an iterative update rule for the backward drift as  
\begin{equation}\label{eq:ipf1}
    \mathbf{b}^{n}_{t+1}(x_{t+1}) = -\mathbf{f}^{n}_{t}(x_{t+1}) + 2\nabla_x \log{p^{f, n}_{t+1}(x_{t+1}}).
\end{equation}

\noindent And the update rule for the forward drift can be derived similarly as
\begin{equation}\label{eq:ipf2}
    \mathbf{f}^{n+1}_{t}(x_t) = -\mathbf{b}^{n}_{t+1}(x_{t}) + 2\nabla_x \log{p^{b, n}_{t}(x_{t}}).
\end{equation}

\noindent With basic calculus and dominated convergence theorem, the followings can be derived:
\begin{equation}\label{eq:derive_obj}
\begin{gathered}
    p^{f,n}(x_{t+1} | x_{t}) = \mathcal{N}(x_{t+1} ; F^{n}_{t}(x_{t}), 2\gamma \mathbf{I}), \\
    p^{f,n}_{t+1}(x_{t+1}) = \mathbf{E}_{p^{f,n}_t}[p^{f,n}(x_{t+1} | x_{t})] \\ 
    = (4\pi \gamma)^{-d/2}\ \mathbf{E}_{p^{f,n}_t}\left[ \exp[{-\Vert F^n_t(x_t) - x_{t+1} \Vert^2}] / 4\gamma \right], \\ 
    \nabla_x \log{p^{f,n}_{t+1}(x_{t+1})} = \mathbf{E}_{p^{f,n}_{t|t+1}}\left[ F^n_t(x_t) - x_{t+1} \right] / 2\gamma, \\
    B^{n}_{t+1}(x_{t+1}) = \mathbf{E}_{p^{f,n}_{t|t+1}}\left[x_{t+1} + F^n_t(x_t) - F^n_{t}(x_{t+1}) \right].
\end{gathered}
\end{equation}

\noindent Based on this result, as proof of other directions is similar, the following iterative objectives can be derived:
\begin{equation}\label{eq:dsb_obj1}
    \mathbf{E}_{p^{f,n}_{t,t+1}}[\Vert B_{t+1}(x_{t+1}) - x_{t+1} - (F^n_{t}(x_t) - F^n_{t}(x_{t+1})) \Vert^2],
\end{equation}
\begin{equation}\label{eq:dsb_obj2}
    \mathbf{E}_{b^{b,n}_{t,t+1}}[\Vert F_{t}(x_{t}) - x_{t} - ( B^{n}_{t+1}(x_{t+1}) - B^{n}_{t+1}(x_{t})) \Vert^2],
\end{equation}

\noindent where $B^{n}_{t+1}$ and $F^{n+1}_{t}$ can be trained with the objectives \ref{eq:dsb_obj1} and \ref{eq:dsb_obj2} respectively.

In the DSB framework, the drift term for Brownian motion cannot be addressed and there is no guarantee that each process $p^f_t$ and $p^b_t$ corresponds to a path measure $p^{\text{SB}}_t$ of SB problem as in SB-FBSDE. However, the non-linear terms $\mathbf{f}_t$ and $\mathbf{b}_t$ are only core drift terms, reducing the need for manual selection of the linear drift terms, unlike FBSDE.
In other words, since the non-linear drifts of DSB are trained through the neural networks without manual selection, the approximation property can be maximized. Thus, the DSB framework can be valid with fewer timesteps compared to SB-FBSDE.
Both SB-FBSDE and DSB can be trained iteratively with Iterative Proportional Fitting (IPF) recursion. And since the number of required iterations for each IPF step is quite large, the training takes a long time for convergence compared to diffusion models.

\section{Proposed Method: Regularization for Schrödinger Bridge}
\label{sec:method}
Based on the SB-based formulations, this study tried to construct discrete-time stochastic processes between any two distributions with smaller timesteps required. While maximizing the use of non-linear drift as in the DSB framework, the formulation of SB-FBSDE was added as a regularization to make the two different forward and backward processes coincide as a path measure of the SB problem.
The existing SB-based generative models' training is unstable with a smaller number of timesteps and iterations. Thus, to construct stable stochastic processes while reducing the number of iterations and timesteps required for convergence, the concept of cycle-consistency proposed by CycleGAN \cite{zhu2017unpaired} was introduced as a regularization to SB-based formulation.

Consider the SB-FBSDE formulation with fixed Brownian motion drift $g_t=1$ and let the both linear and non-linear drift of SB-FBSDE as a single non-linear drift as in DSB,
\begin{equation}\label{eq:new_f}
    dx_t = \left[ f_{t}(x_t) + \nabla_x \log{\psi_t(x_t)} \right]dt + dw_t = \mathbf{f}_{t}(x_t)dt + dw_t,
\end{equation}
\begin{equation}\label{eq:new_b}
    dx_t = \left[ f_{t}(x_t) - \nabla_x \log{\hat{\psi}_t(x_t)} \right]dt + dw_t = - \mathbf{b}_{t}(x_t)dt + dw_t.
\end{equation}

\noindent When considering the following forward and backward processes where only the degree of variance differs from \ref{eq:dsb_f} and \ref{eq:dsb_b} of DSB,
\begin{equation}\label{eq:regsb_f}
\begin{gathered}
    p^f(x_{t+1} | x_{t}) = \mathcal{N}(x_{t+1}; x_{t} + \gamma \mathbf{f}_{t}(x_t), \gamma \mathbf{I}) 
    = \mathcal{N}(x_{t+1} ; F_{t}(x_{t}), \gamma \mathbf{I}), \\
    \rightarrow\quad x_{t+1} = x_{t} + \gamma \mathbf{f}_{t}(x_t) + \sqrt{\gamma} z
\end{gathered}
\end{equation}
\begin{equation}\label{eq:regsb_b}
\begin{gathered}
    p^b(x_{t} | x_{t+1}) = \mathcal{N}(x_{t}; x_{t+1} + \gamma \mathbf{b}_{t+1}(x_{t+1}), \gamma \mathbf{I})
    = \mathcal{N}(x_t ; B_{t+1}(x_{t+1}), \gamma \mathbf{I}), \\
    \rightarrow\quad x_{t} = x_{t+1} - \gamma \left(-\mathbf{b}_{t+1}(x_{t+1}) \right) + \sqrt{\gamma} z.
\end{gathered}
\end{equation}

\noindent Note that with different degree of variance, the \ref{eq:regsb_f} and \ref{eq:regsb_b} can have the same objective functions \ref{eq:dsb_obj1} and \ref{eq:dsb_obj2} respectively through the same process of DSB.
Interpreting the above discrete stochastic processes as the Euler-Maruyama scheme with discretization step size $\gamma$, the corresponding continuous-time SDEs with forward and backward directions are given as
\begin{equation}\label{eq:regsb_f_sde}
    dx_t = \mathbf{f}_{t}(x_t)dt + dw_t,
\end{equation}
\begin{equation}\label{eq:regsb_b_sde}
    dx_t = - \mathbf{b}_{t}(x_t)dt + dw_t,
\end{equation}

\noindent Since these expressions \ref{eq:regsb_f_sde} and \ref{eq:regsb_b_sde} are equivalent to \ref{eq:new_f} and \ref{eq:new_b} respectively, it demonstrates that DSB and SB-FBSDE can be related as continuous-time SDEs. Note that with the property of SB-FBSDE formulation \ref{eq:fbsde_prop}, the following holds:
\begin{equation}\label{eq:newSB_derive1}
    \mathbf{f}_{t}(x_t) + \mathbf{b}_{t}(x_t) = \nabla_x \log{p^{\text{SB}}_{t}(x_t)}.
\end{equation}

\noindent And in ideal scenario, the path measure $p^{\text{SB}}_{t}$ of SB problem should coincide with marginal measure of the forward and backward processes, \ref{eq:regsb_f_sde} and \ref{eq:regsb_b_sde}. In this case, the following relationships hold:
\begin{equation}\label{eq:newSB_derive3}
    \nabla_x \log{p^{f}_{t}(x_t)} = \nabla_x \log{p^{\text{SB}}_{t}(x_t)} = \nabla_x \log{p^{b}_{t}(x_t)},
\end{equation}

\noindent where $p^{f}_{t}$ and $p^{b}_{t}$ corresponds to the marginal distribution of forward and backward processes, \ref{eq:regsb_f_sde} and \ref{eq:regsb_b_sde} respectively. Now, through the same process as the work of DSB, the following holds:
\begin{equation}\label{eq:newSB_derive2}
    \nabla_x \log{p^{f,n}_{t+1}(x_{t+1})} = \mathbf{E}_{p^{f,n}_{t|t+1}}\left[ F^n_t(x_t) - x_{t+1} \right] / \gamma.
\end{equation}

\noindent Combining the results of \ref{eq:newSB_derive1}, \ref{eq:newSB_derive2}, and \ref{eq:newSB_derive2}, the followings can be derived:
\begin{equation}\label{eq:derive_4}
\begin{gathered}
    \nabla_x \log{p^{\text{SB}, n}_{t+1}(x_{t+1})} = \nabla_x \log{p^{f, n}_{t+1}(x_{t+1})}, \\
    \mathbf{f}^n_{t+1}(x_{t+1}) + \mathbf{b}^n_{t+1}(x_{t+1}) = \mathbf{E}_{p^{f,n}_{t|t+1}}\left[ F^n_t(x_t) - x_{t+1} \right] / \gamma, \\
    \gamma \mathbf{b}^{n+1}_{t+1}(x_{t+1}) = \mathbf{E}_{p^{f,n}_{t|t+1}}\left[ F^n_t(x_t) - F^n_{t+1}(x_{t+1}) \right], \\
    B^{n}_{t+1}(x_{t+1}) = \mathbf{E}_{p^{f,n}_{t|t+1}}\left[x_{t+1} + F^n_t(x_t) - F^n_{t+1}(x_{t+1}) \right].
\end{gathered}
\end{equation}

\noindent Based on this, as a case of other directions is similar, the following regularization objectives are given as
\begin{equation}\label{eq:reg1}
    \mathbf{E}_{p^{f,n}_{t,t+1}}[\Vert B_{t+1}(x_{t+1}) - x_{t+1} - (F^n_{t}(x_t) - F^n_{t+1}(x_{t+1})) \Vert^2],
\end{equation}
\begin{equation}\label{eq:reg2}
    \mathbf{E}_{b^{b,n}_{t,t+1}}[\Vert F_{t}(x_{t}) - x_{t} - ( B^{n}_{t+1}(x_{t+1}) - B^{n}_{t}(x_{t})) \Vert^2],
\end{equation}

\noindent where \ref{eq:reg1} holds for $t \in \{0,\dots,T-2 \}$ and \ref{eq:reg2} holds for $t \in \{1,\dots,T-1 \}$. And $B^{n}_{t+1}$ and $F^{n+1}_{t}$ can be iteratively trained with the objectives \ref{eq:reg1} and \ref{eq:reg2} respectively.
This objective can be thought as an additional regularization term for the DSB's objective function \ref{eq:dsb_obj1} and \ref{eq:dsb_obj2} since it cannot applied for all time step $t \in \{0,\dots,T-1 \}$.

Putting objective term of \ref{eq:dsb_obj1} as $\mathcal{L}_{DSB}$ and \ref{eq:reg1} as $\mathcal{L}_{reg}$, the objective of regularized SB-based formulation can be expressed as 
\begin{equation}\label{eq:sb_reg}
    \alpha \mathcal{L}_{DSB} + (1 - \alpha) \mathcal{L}_{reg},
\end{equation}

\noindent where $\alpha$ is a hyperparameter. Note that the $\mathcal{L}_{DSB}$ and $\mathcal{L}_{reg}$ are very similar, and learning these two objectives as separate terms requires more GPU memory by storing two similar computational graphs. With the convexity of $\Vert \cdot \Vert^2$, the memory-efficient objective can be attained as
\begin{equation}\label{eq:obj_reg}
\begin{gathered}
    \alpha \mathcal{L}_{DSB} + (1 - \alpha) \mathcal{L}_{reg} \geq 
    \mathcal{L}_{memory} = \\
    \mathbf{E}_{p^{f,n}_{t, t+1}}[\Vert B(x_{t+1}) - x_{t+1} - (F^n_{t}(x_t) - \alpha F^n_{t}(x_{t+1}) - (1 - \alpha) F^n_{t+1}(x_{t+1})) \Vert^2].
\end{gathered}
\end{equation}

Now, note that stable forward and backward processes in an ideal scenario should satisfy the following
\begin{equation}\label{eq:cyc}
    B_{t+1}(F_t(x_{t})) = x_t, \quad F_{t}(B_{t+1}(x_{t+1})) = x_{t+1}, \quad t \in \{0,\dots,T-1 \}.
\end{equation}

\noindent Note that the above relation can be considered as a cycle-consistency constraint proposed by CycleGAN \cite{zhu2017unpaired}. It is essential relation that must hold for entire stochastic processes, but SB-based processes with small timesteps may not attain this.
Therefore, to reduce the number of iterations and timesteps required for convergence while maintaining the stability of the constructed stochastic process, the above cycle-consistency relation can be used as an additional regularization term to the objective function explicitly.
For this, the above relation \ref{eq:cyc} can be expressed as
\begin{equation}\label{eq:cyc1}
\begin{gathered}
    \mathbf{E}_{x_{t+1} \sim\ p^f_{t+1 | t}}\left[B_{t+1}(x_{t+1}) \right] = x_t, \quad
    \mathbf{E}_{x_{t} \sim\ p^b_{t | t+1}}\left[F_{t}(x_{t}) \right] = x_{t+1}, \\
    t \in \{0,\dots,T-1 \}.
\end{gathered}
\end{equation}

\noindent And again, it can be formulated as the following regression problem:
\begin{equation}\label{eq:cyc2}
    \mathcal{L}_{cyc} = \mathbf{E}_{p^{f,n}_{t, t+1}}[\Vert B_{t+1}(x_{t+1}) - x_{t} \Vert^2].
\end{equation}

\noindent This objective, $\mathcal{L}_{cyc}$, is an additional term of $\mathcal{L}_{memory}$. Note that all objectives $\mathcal{L}_{DSB}$, $\mathcal{L}_{reg}$, and $\mathcal{L}_{cyc}$ have $B_{t+1}(x_{t+1})$ term, this is the only term that is evaluated during training. Thus, for efficient training, $B_{t+1}(x_{t+1})$ should be evaluated once. 
If $\mathcal{L}_{DSB}$, $\mathcal{L}_{reg}$, and $\mathcal{L}_{cyc}$ terms are used separately, evaluation of $B_{t+1}(x_{t+1})$ should occur multiple times, which is inefficient. 
By the convexity of $\Vert \cdot \Vert^2$ and with the proper setting of the weight $\beta$, the memory-efficient loss objective can be obtained as 
\begin{equation}\label{eq:loss_b}
\begin{aligned}
    \mathcal{L}_{B} = \mathbf{E}_{p^{f,n}_{t, t+1}} \left[ \Vert B_{t+1}(x_{t+1}) - \frac{1}{\beta + 1} \left( x_{t+1} - (F^n_{t}(x_t) - \alpha F^n_{t}(x_{t+1}) - \right. \right. \\ \left. \left.
    (1 - \alpha) F^n_{t+1}(x_{t+1})) \right) - \frac{\beta}{\beta + 1} x_{t} \Vert^2 \right] \leq \frac{1}{\beta + 1} \mathcal{L}_{memory} + \frac{\beta}{\beta + 1} \mathcal{L}_{cyc}.
\end{aligned}
\end{equation}

\noindent Similarly, the loss objective for updating $F_{t}^{n+1}$ can be obtained as
\begin{equation}\label{eq:loss_f}
\begin{aligned}
    \mathcal{L}_{F} = \mathbf{E}_{p^{b,n}_{t, t+1}} \left[ \Vert F_{t}(x_{t}) - \frac{1}{\beta + 1} \left( x_{t} - (B^n_{t+1}(x_{t+1}) - \alpha B^n_{t+1}(x_{t}) - \right. \right. \\ \left. \left.
    (1 - \alpha) B^n_{t}(x_{t})) \right) - \frac{\beta}{\beta + 1} x_{t+1} \Vert^2 \right].
\end{aligned}
\end{equation}

\noindent Thus, the regularized SB-based model, \textbf{RSB}, is trained for the desired bidirectional stochastic processes by alternating between $\mathcal{L}_{B}$ and $\mathcal{L}_{F}$. Since the objective was set to memory-efficient style, model evaluation $B_{t+1}(x_{t+1})$ or $F_{t}(x_{t})$ proceeds only once for each update. The remaining term of the objectives can be effectively obtained by a replay-memory \cite{mnih2013playing}.

\section{Experiments}
\label{sec:experiments}
Based on the regularized SB-based formulation, the proposed RSB experimentally demonstrated that it can train stochastic processes between any two data spaces with relatively small timesteps compared to the previous SB models. And its training was more stable and faster.
Since the SB-based stochastic process is free from the constraints of starting at $\mathcal{Z}$, the experiments were conducted on both unconditional and conditional generation tasks. And for both types of tasks, the proposed RSB confirmed its effectiveness.

\subsection{Dataset \& Training}
2D Toy, MNIST, and CelebA were used as datasets for qualitative performance evaluation of RSB.
The 2D Toy dataset consists of intuitive 2-dimensional data, including 8-Gaussian, Checkerboard, 25-Gaussian, and Circles.
And the MNIST is one of the most widely known datasets in deep learning research and consists of digits ranging from 0 to 9. 
Lastly, the CelebA dataset, where various male and female faces exist at various ages, was also used to determine whether the RSB can handle the relatively high-resolution image domain.

A vanilla GAN model with a gradient penalty (GP) of the form \cite{mescheder2018training} was trained on the 2D Toy to compare with the proposed RSB. And for other datasets except for the 2D Toy dataset, the NCSN++ architecture of SGM \cite{song2020score} was used. 
In addition, since RSB requires multiple steps of model evaluation, the training can be very slow if the model is evaluated for each update iteration. To mitigate this issue, the replay memory was generated by inferencing with a large batch size at once and the generated replay memory was used for multiple iterations.
Also, for faster training, in the case of unconditional generation, since the forward noising process from $\mathcal{P}$ to $\mathcal{Z}$ is relatively easy to be trained, half of the number of iterations was used at each IPF stage compared to training of the backward process.

\subsection{Results}
The existing SB models, DSB and SB-FBSDE, can be compared to RSB for performance evaluation. Since SB-FBSDE depends heavily on continuous-time SDEs, SB-FBSDE requires 100 steps even for a 2D Toy dataset. Thus it could not be trained for small timesteps such as $T=4,8$. And since DSB was formulated in a relatively discrete-time setting compared to SB-FBSDE, DSB showed better performance when the number of timesteps is restricted to be small. Thus, the proposed RSB was compared to DSB.

\subsubsection{Results with 2D Toy}
\begin{figure*}
\begin{center}
\centering
    \begin{subfigure}{.32\textwidth}
    \includegraphics[width=\linewidth]{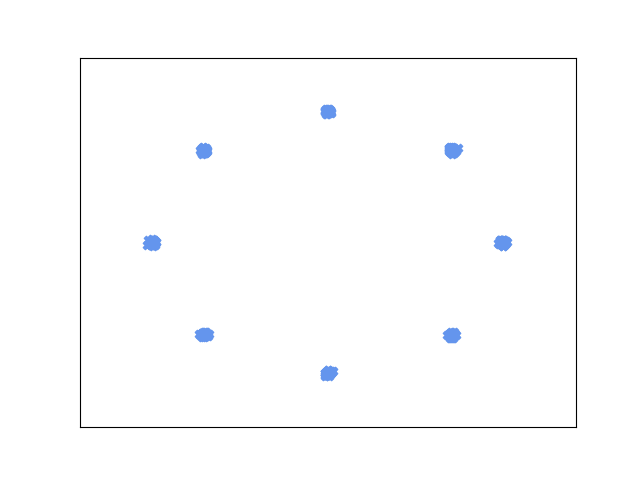}
    \vspace*{-7.5mm}
    \caption{Ground Truth}
    \end{subfigure}
    \begin{subfigure}{.32\textwidth}
    \includegraphics[width=\linewidth]{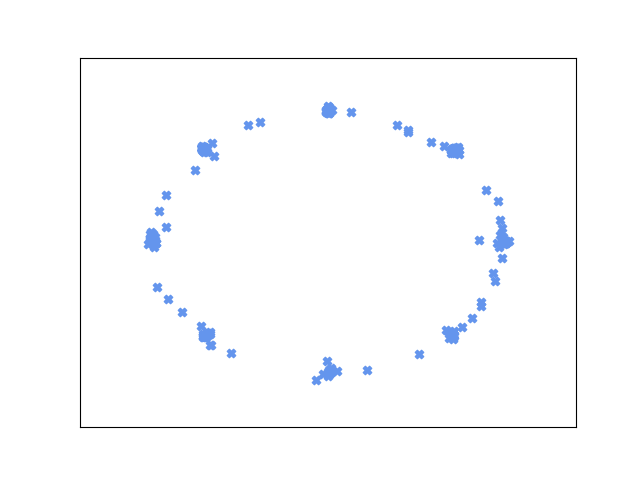}
    \vspace*{-7.5mm}
    \caption{DSB}
    \end{subfigure}
    \begin{subfigure}{.32\textwidth}
    \includegraphics[width=\linewidth]{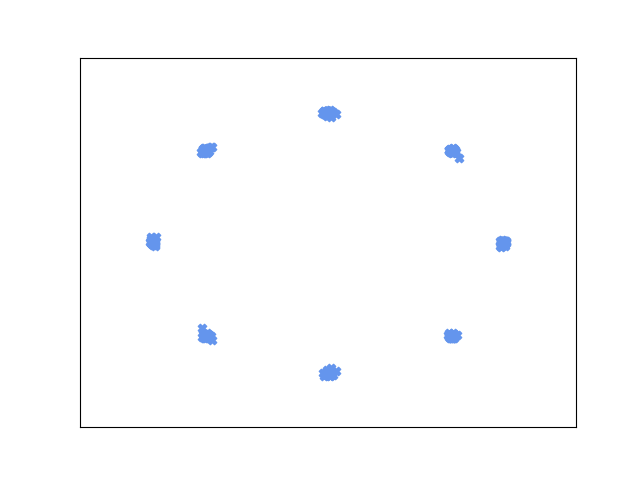}
    \vspace*{-7.5mm}
    \caption{\textbf{RSB}}
    \end{subfigure}
\caption{Qualitative results on 8-Gaussians of 2D Toy}
\label{fig:toy_8gauss}
\end{center}
\end{figure*}

Firstly, RSB was trained for unconditional generation of 8-Gaussians of 2D Toy. Both RSB and DSB were trained with 8 timesteps, 10K iterations for the forward process, and 20K iterations for the backward process. And hyperparameters for RSB was set to $\alpha=0.5$ and $\beta=2.5$. See Figure \ref{fig:toy_8gauss} for comparison. While DSB didn't converge to the desired data space, RSB almost converged.

\begin{figure*}
\begin{center}
\centering
    \begin{subfigure}{.32\textwidth}
    \includegraphics[width=\linewidth]{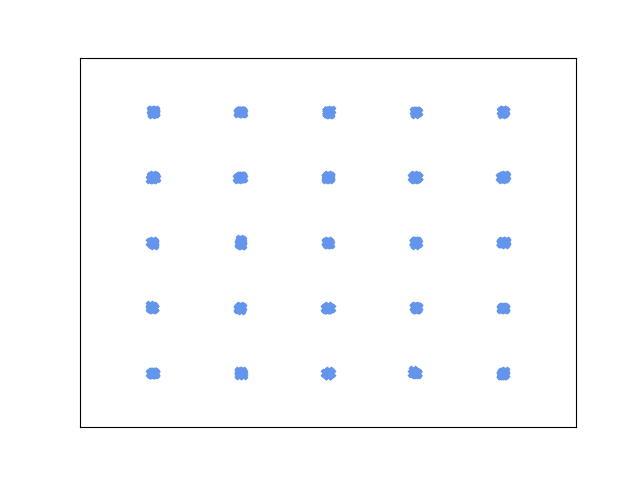}
    \vspace*{-7.5mm}
    \caption{Ground Truth}
    \end{subfigure}
    \begin{subfigure}{.32\textwidth}
    \includegraphics[width=\linewidth]{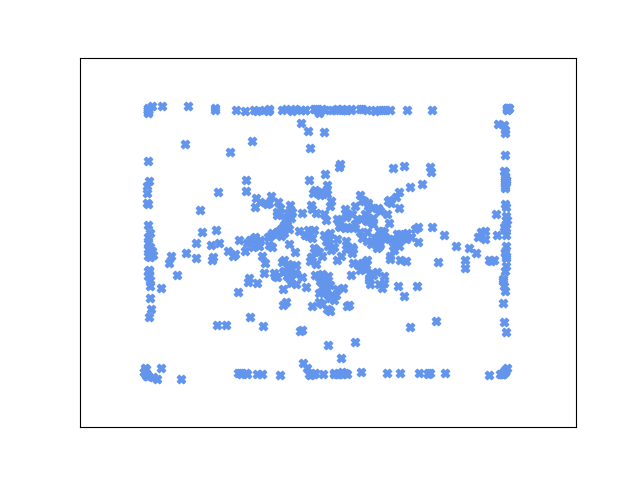}
    \vspace*{-7.5mm}
    \caption{DSB(intermediate)}
    \end{subfigure}
    \begin{subfigure}{.32\textwidth}
    \includegraphics[width=\linewidth]{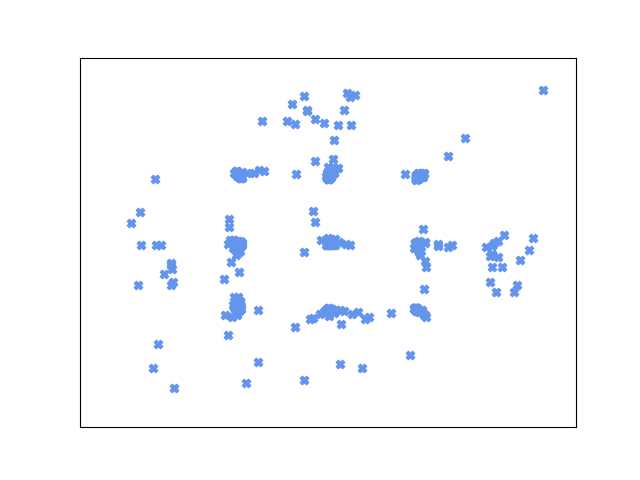}
    \vspace*{-7.5mm}
    \caption{DSB}
    \end{subfigure}
    \begin{subfigure}{.32\textwidth}
    \includegraphics[width=\linewidth]{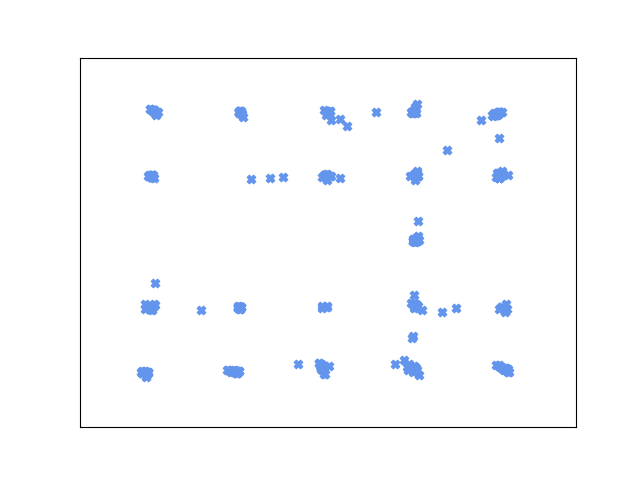}
    \vspace*{-7.5mm}
    \caption{GAN}
    \end{subfigure}
    \begin{subfigure}{.32\textwidth}
    \includegraphics[width=\linewidth]{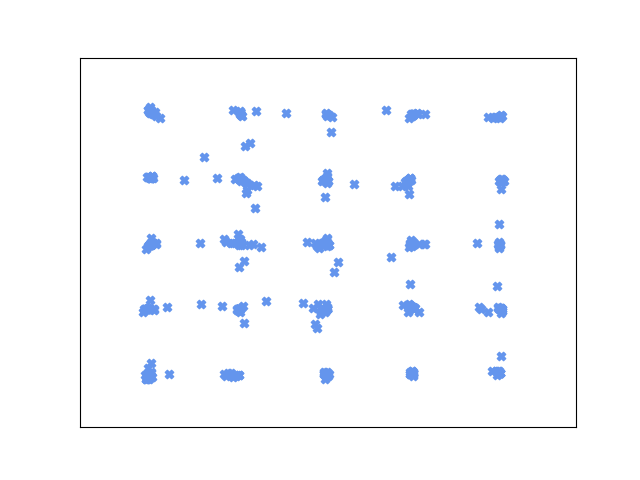}
    \vspace*{-7.5mm}
    \caption{\textbf{RSB}(intermediate)}
    \end{subfigure}
    \begin{subfigure}{.32\textwidth}
    \includegraphics[width=\linewidth]{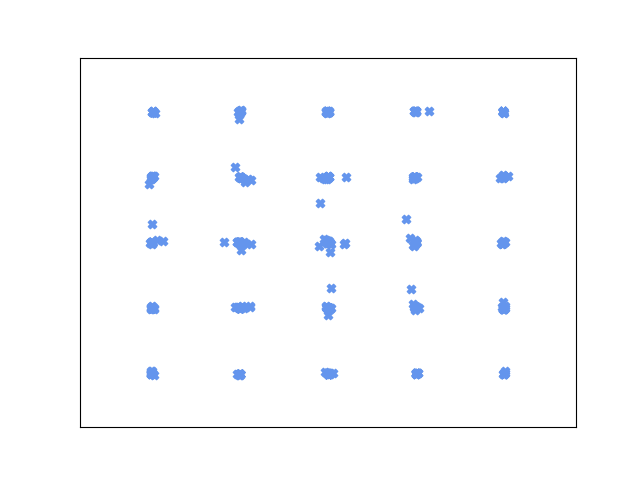}
    \vspace*{-7.5mm}
    \caption{\textbf{RSB}}
    \end{subfigure}
\caption{Qualitative results on 25-Gaussians of 2D Toy}
\label{fig:toy_25gauss}
\end{center}
\end{figure*}

Next, an unconditional generation of 25-Gaussians which is much more complicated than 8-Gaussians was testified. Both RSB and DSB were trained with 8 timesteps, 20K iterations for the forward process, and 40K iterations for the backward process. And hyperparameters for RSB was set to $\alpha=0.5$ and $\beta=5$. And GAN was trained for 60K iterations.
See Figure \ref{fig:toy_25gauss} for comparison between DSB, RSB and GAN. While DSB showed instability where the training did not progress significantly after the intermediate stage, RSB showed fast convergence in the intermediate stage, and the rest of the training stayed stable. 
This indicates that the existing SB-based methods are not suitable for a small number of discretized timesteps while RSB is. And although GAN was trained for a sufficient training time, mode collapse occurred that it could not cover all modes of 25-Gaussians. The result shows that stochastic-process-based generative modeling can complement the existing function-based one.

\begin{figure*}
\begin{center}
\centering
    \begin{subfigure}{.24\textwidth}
    \includegraphics[width=\linewidth]{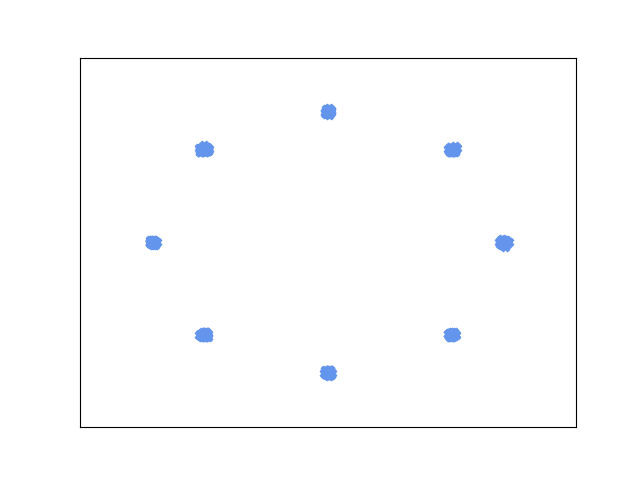}
    
    \vspace{-.4cm}
    
    \includegraphics[width=\linewidth]{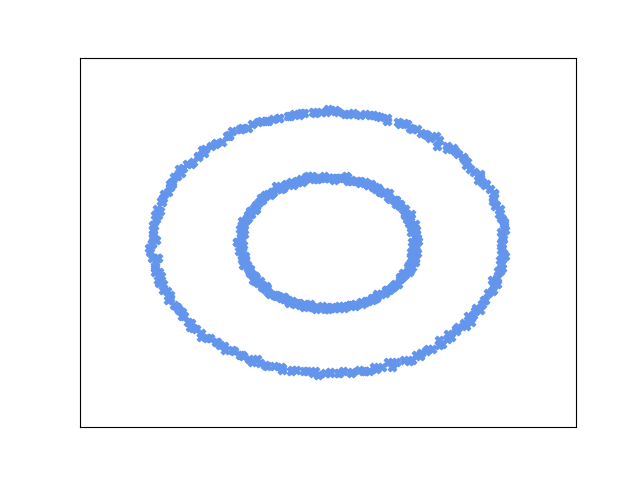}
    \vspace*{-7.5mm}
    \caption{Ground Truth}
    \end{subfigure}
    \begin{subfigure}{.24\textwidth}
    \includegraphics[width=\linewidth]{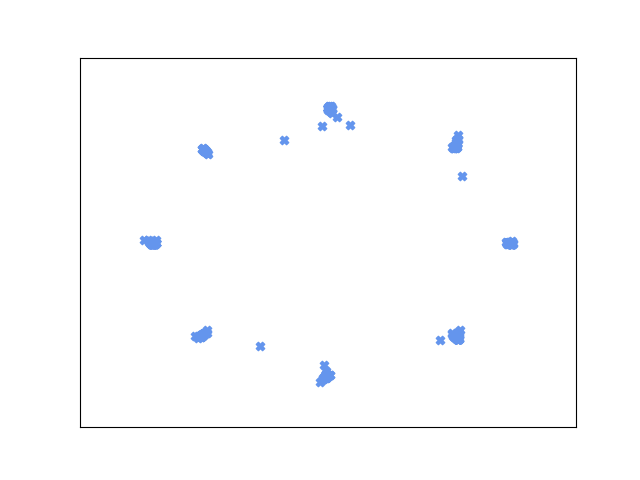}
    
    \vspace{-.4cm}
    
    \includegraphics[width=\linewidth]{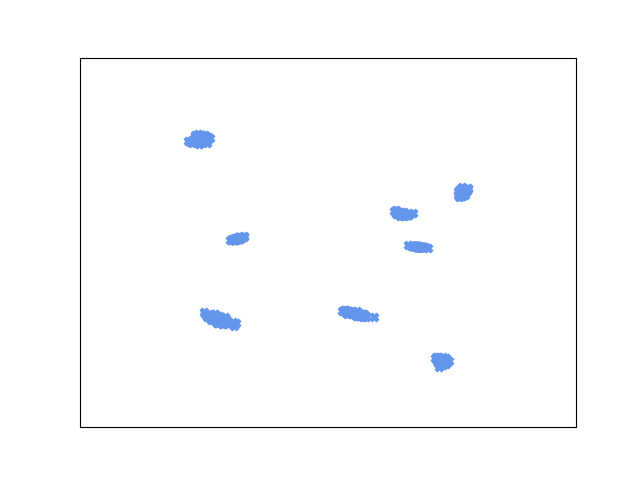}
    \vspace*{-7.5mm}
    \caption{GAN}
    \end{subfigure}
    \begin{subfigure}{.24\textwidth}
    \includegraphics[width=\linewidth]{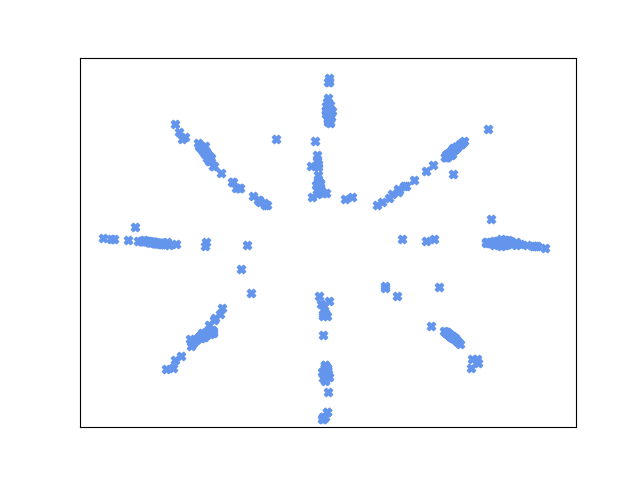}
    
    \vspace{-.4cm}
    
    \includegraphics[width=\linewidth]{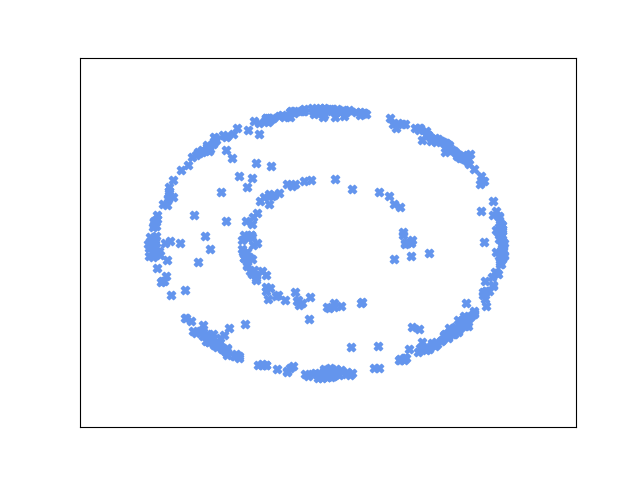}
    \vspace*{-7.5mm}
    \caption{DSB}
    \end{subfigure}
    \begin{subfigure}{.24\textwidth}
    \includegraphics[width=\linewidth]{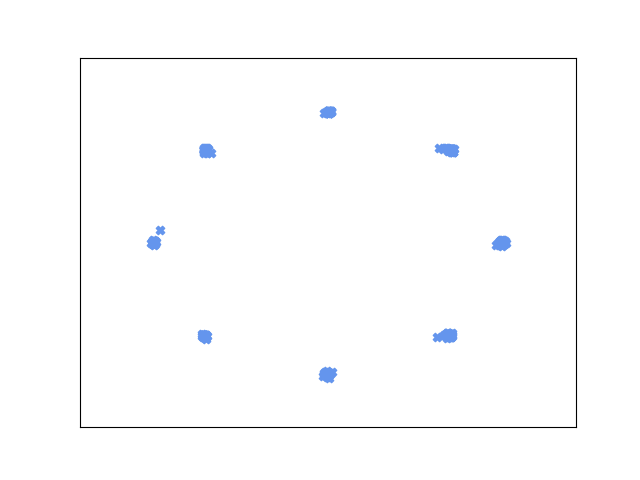}
    
    \vspace{-.4cm}
    
    \includegraphics[width=\linewidth]{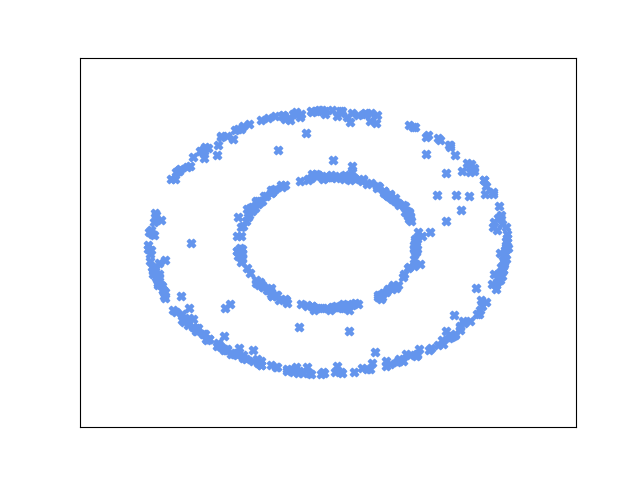}
    \vspace*{-7.5mm}
    \caption{\textbf{RSB}}
    \end{subfigure}
\caption{Qualitative results on data translation task between 8-Gaussians and Circles of 2D Toy}
\label{fig:toy_8gauss2circ}
\end{center}
\end{figure*}

\begin{figure*}
\begin{center}
\centering
    \begin{subfigure}{.19\textwidth}
    \includegraphics[width=\linewidth]{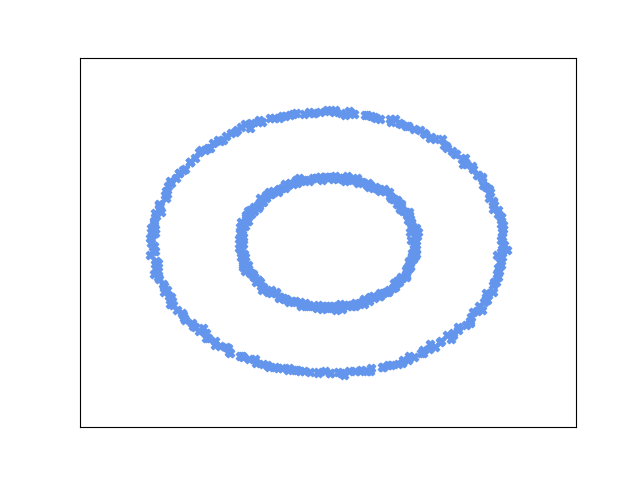}
    \end{subfigure}
    \begin{subfigure}{.19\textwidth}
    \includegraphics[width=\linewidth]{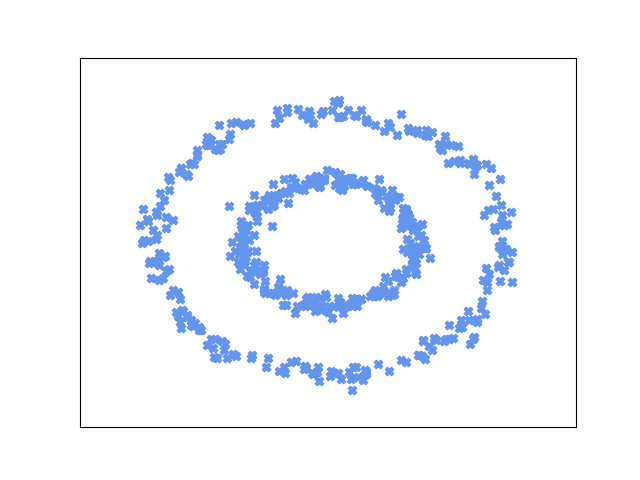}
    \end{subfigure}
    \begin{subfigure}{.19\textwidth}
    \includegraphics[width=\linewidth]{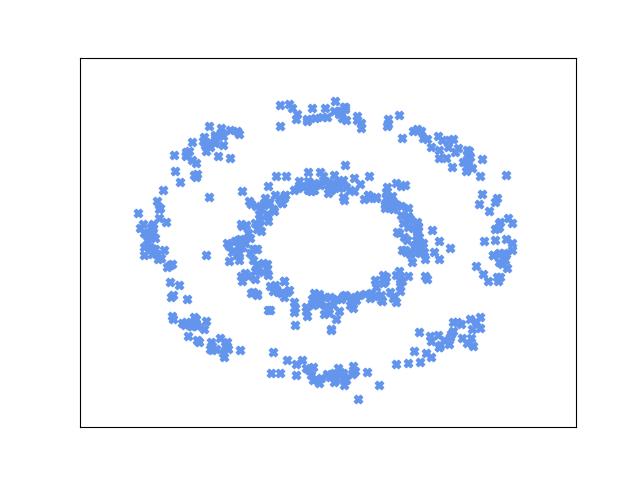}
    \end{subfigure}
    \begin{subfigure}{.19\textwidth}
    \includegraphics[width=\linewidth]{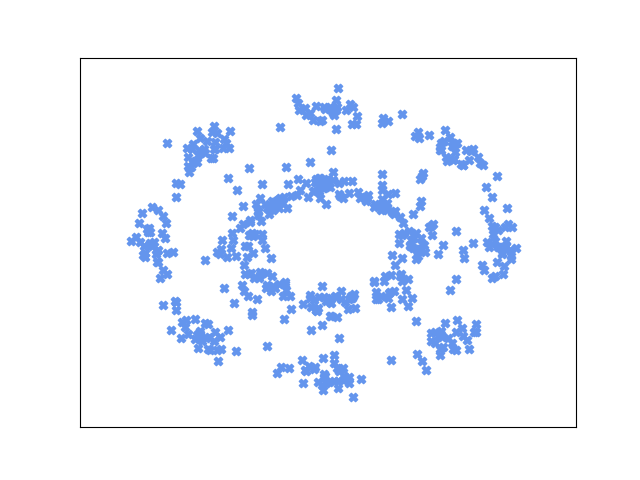}
    \end{subfigure}
    \begin{subfigure}{.19\textwidth}
    \includegraphics[width=\linewidth]{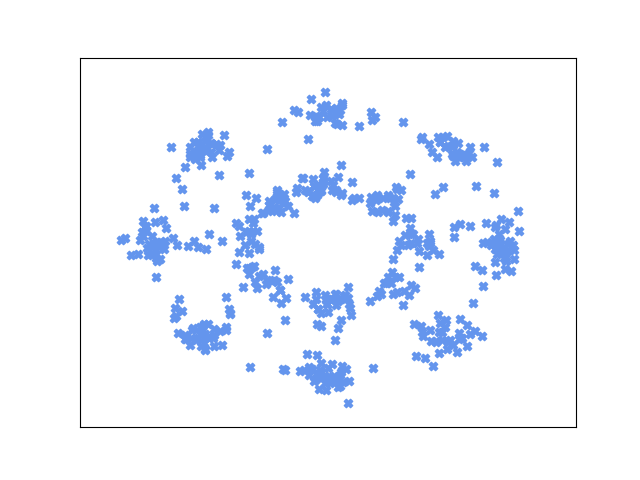}
    \end{subfigure}
    
    \begin{subfigure}{.19\textwidth}
    \includegraphics[width=\linewidth]{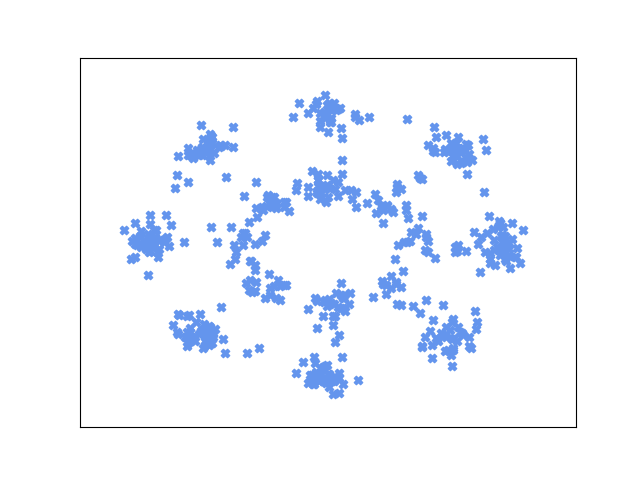}
    \end{subfigure}
    \begin{subfigure}{.19\textwidth}
    \includegraphics[width=\linewidth]{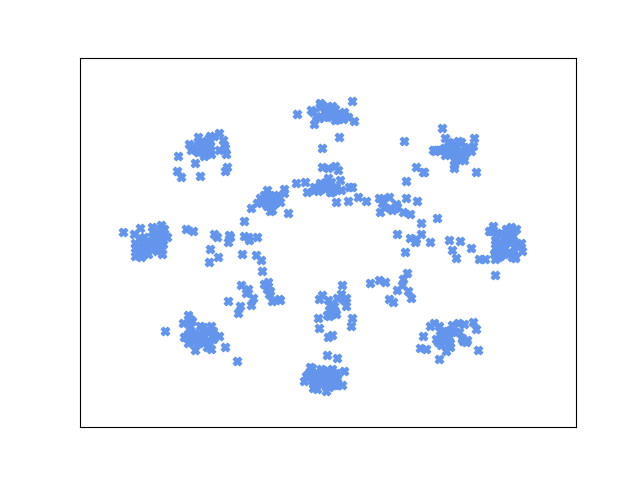}
    \end{subfigure}
    \begin{subfigure}{.19\textwidth}
    \includegraphics[width=\linewidth]{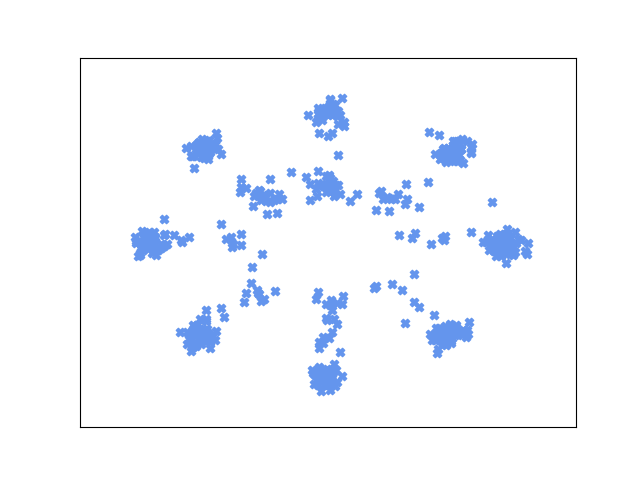}
    \end{subfigure}
    \begin{subfigure}{.19\textwidth}
    \includegraphics[width=\linewidth]{fig/toy/8gauss2circ/dsb/f_8.png}
    \end{subfigure}
    
    \begin{subfigure}{.19\textwidth}
    \includegraphics[width=\linewidth]{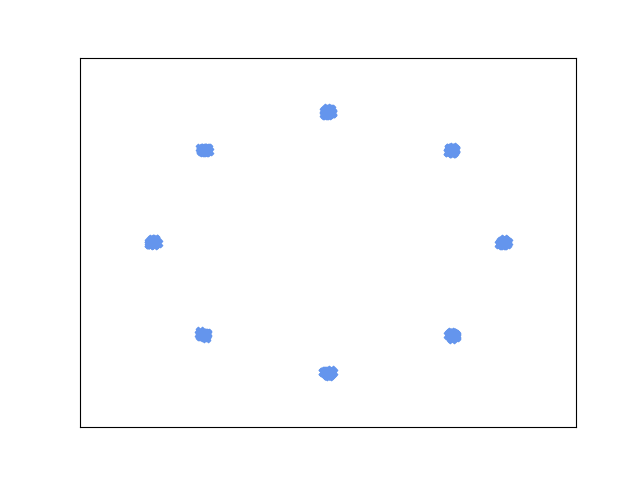}
    \end{subfigure}
    \begin{subfigure}{.19\textwidth}
    \includegraphics[width=\linewidth]{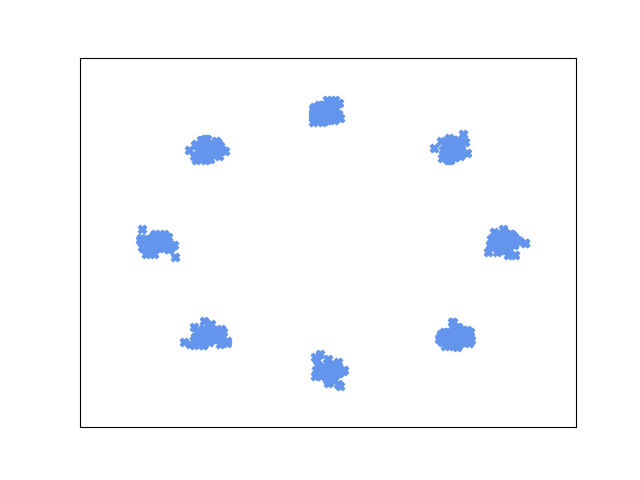}
    \end{subfigure}
    \begin{subfigure}{.19\textwidth}
    \includegraphics[width=\linewidth]{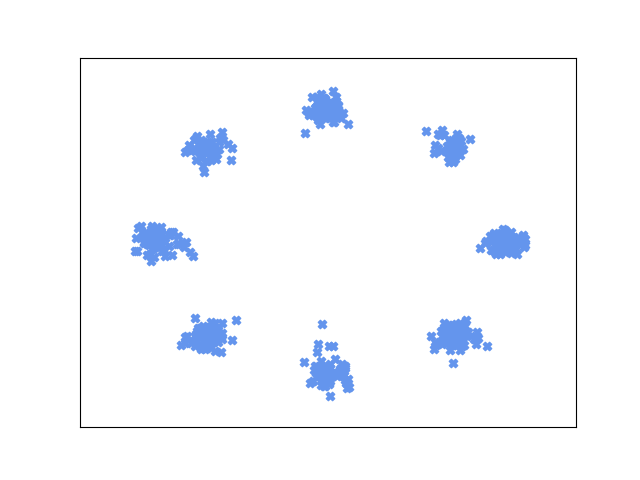}
    \end{subfigure}
    \begin{subfigure}{.19\textwidth}
    \includegraphics[width=\linewidth]{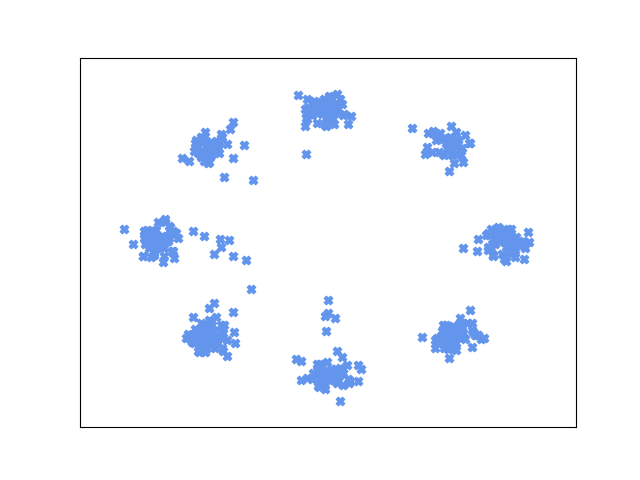}
    \end{subfigure}
    \begin{subfigure}{.19\textwidth}
    \includegraphics[width=\linewidth]{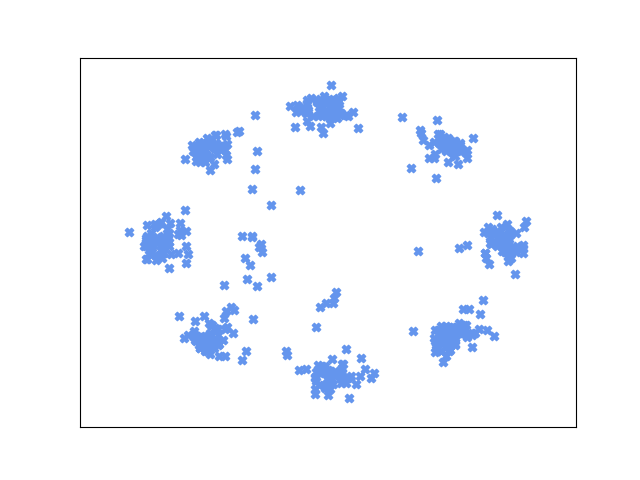}
    \end{subfigure}
    
    \begin{subfigure}{.19\textwidth}
    \includegraphics[width=\linewidth]{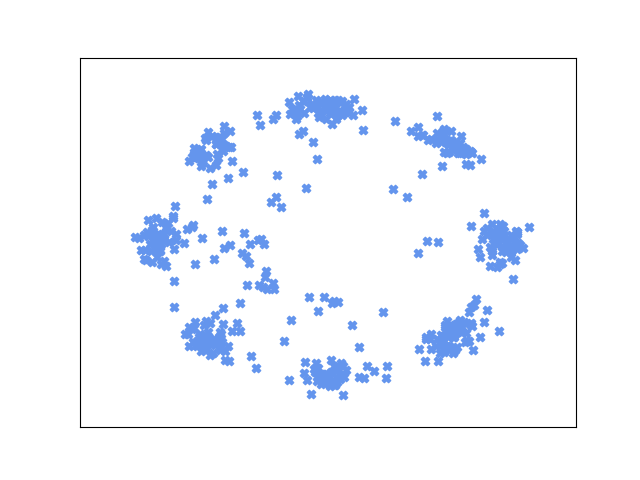}
    \end{subfigure}
    \begin{subfigure}{.19\textwidth}
    \includegraphics[width=\linewidth]{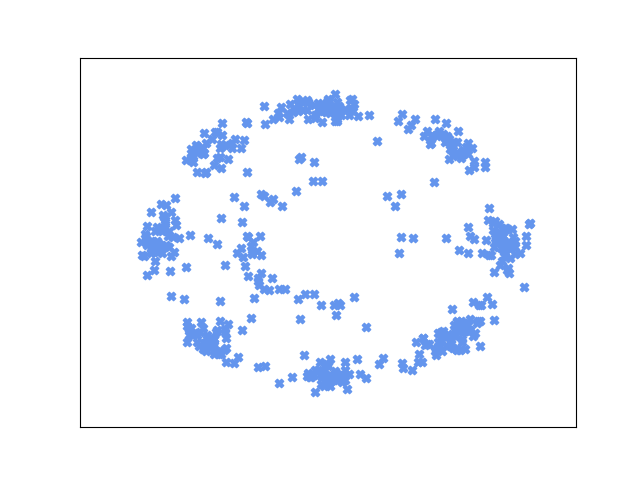}
    \end{subfigure}
    \begin{subfigure}{.19\textwidth}
    \includegraphics[width=\linewidth]{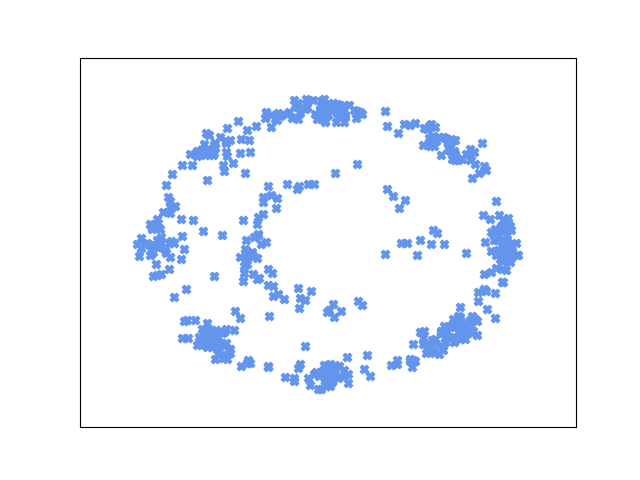}
    \end{subfigure}
    \begin{subfigure}{.19\textwidth}
    \includegraphics[width=\linewidth]{fig/toy/8gauss2circ/dsb/b_8.png}
    \end{subfigure}
\caption{Detailed qualitative results of DSB on data translation between 8-Gaussians and Circles of 2D Toy. The top two rows illustrate the forward process from Circles to 8-Gaussians. And other rows illustrate the backward process from 8-Gaussians to Circles.}
\label{fig:toy_8gauss2circ_dsb}
\end{center}
\end{figure*}

\begin{figure*}
\begin{center}
\centering
    \begin{subfigure}{.19\textwidth}
    \includegraphics[width=\linewidth]{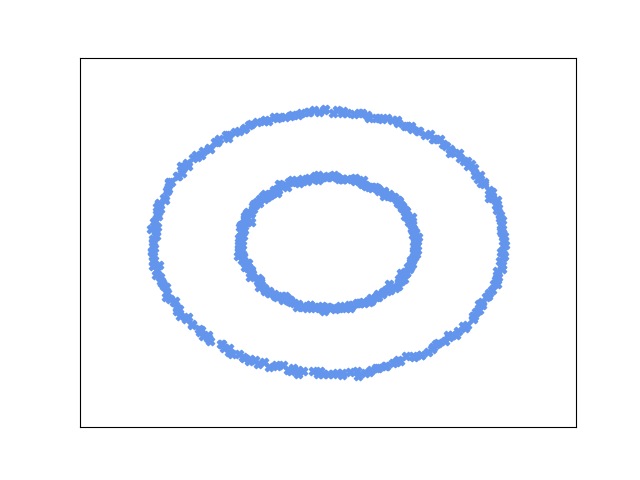}
    \end{subfigure}
    \begin{subfigure}{.19\textwidth}
    \includegraphics[width=\linewidth]{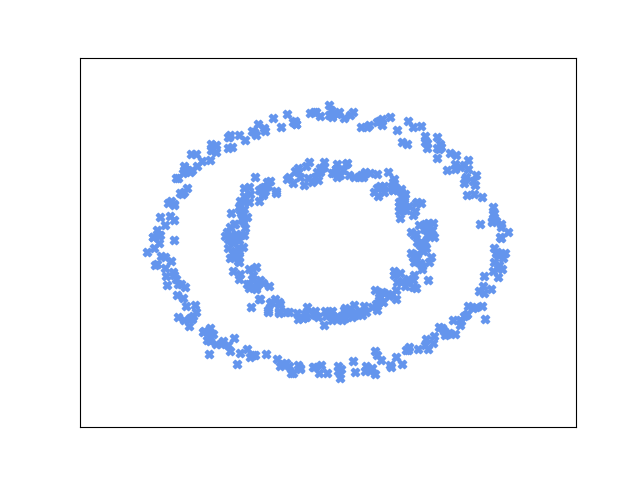}
    \end{subfigure}
    \begin{subfigure}{.19\textwidth}
    \includegraphics[width=\linewidth]{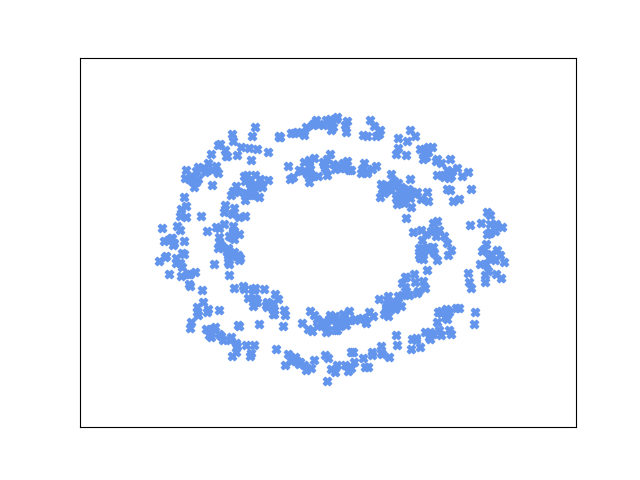}
    \end{subfigure}
    \begin{subfigure}{.19\textwidth}
    \includegraphics[width=\linewidth]{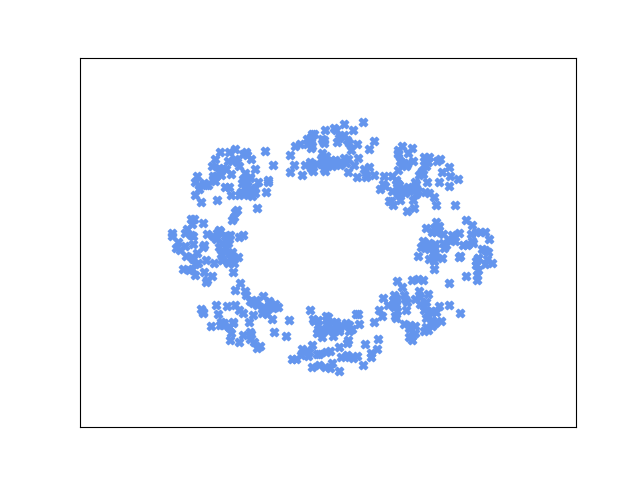}
    \end{subfigure}
    \begin{subfigure}{.19\textwidth}
    \includegraphics[width=\linewidth]{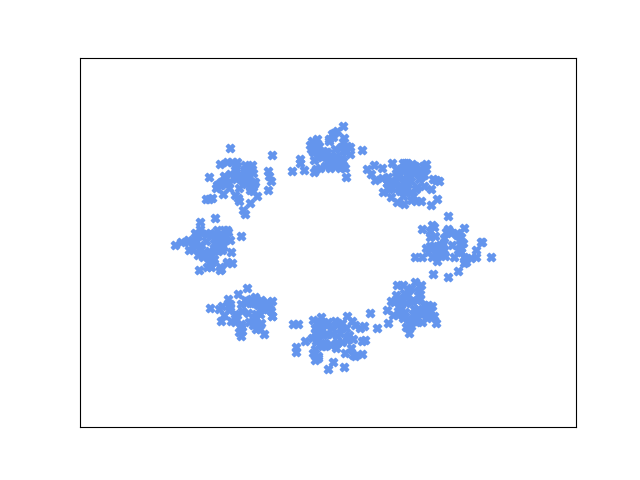}
    \end{subfigure}
    
    \begin{subfigure}{.19\textwidth}
    \includegraphics[width=\linewidth]{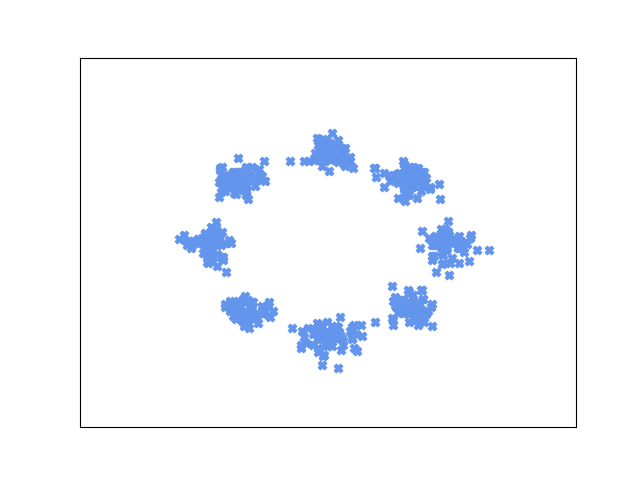}
    \end{subfigure}
    \begin{subfigure}{.19\textwidth}
    \includegraphics[width=\linewidth]{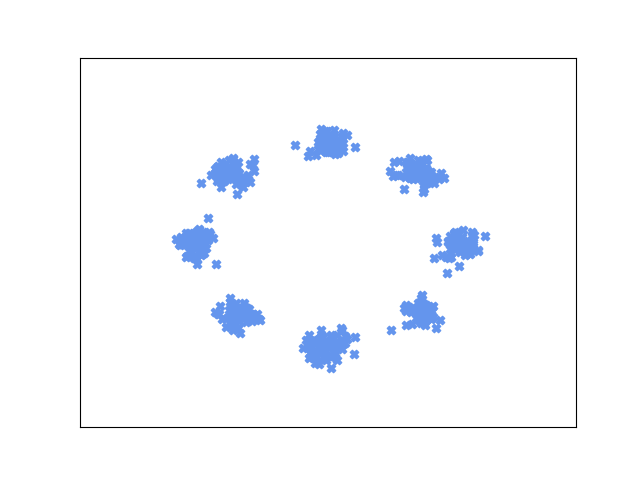}
    \end{subfigure}
    \begin{subfigure}{.19\textwidth}
    \includegraphics[width=\linewidth]{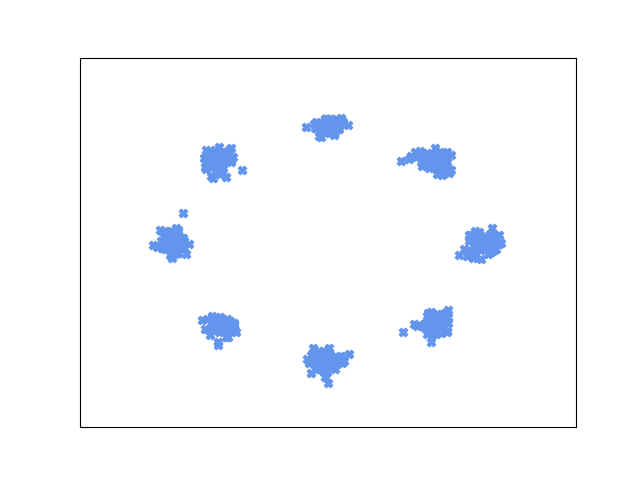}
    \end{subfigure}
    \begin{subfigure}{.19\textwidth}
    \includegraphics[width=\linewidth]{fig/toy/8gauss2circ/rsb/f_8.png}
    \end{subfigure}
    
    \begin{subfigure}{.19\textwidth}
    \includegraphics[width=\linewidth]{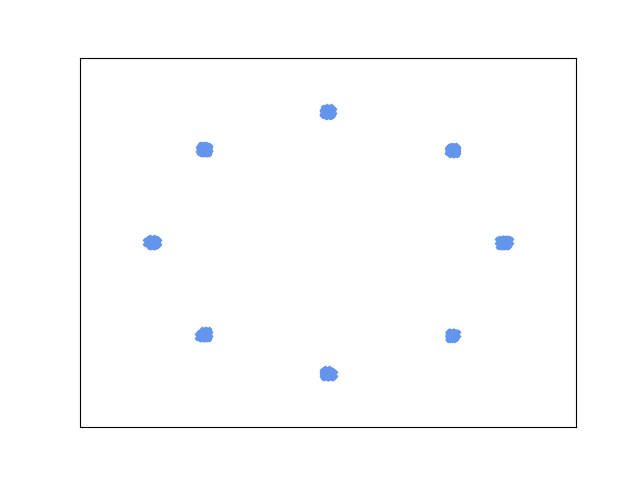}
    \end{subfigure}
    \begin{subfigure}{.19\textwidth}
    \includegraphics[width=\linewidth]{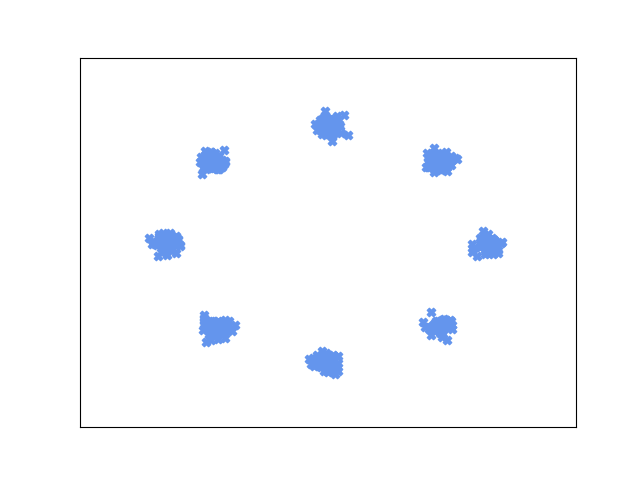}
    \end{subfigure}
    \begin{subfigure}{.19\textwidth}
    \includegraphics[width=\linewidth]{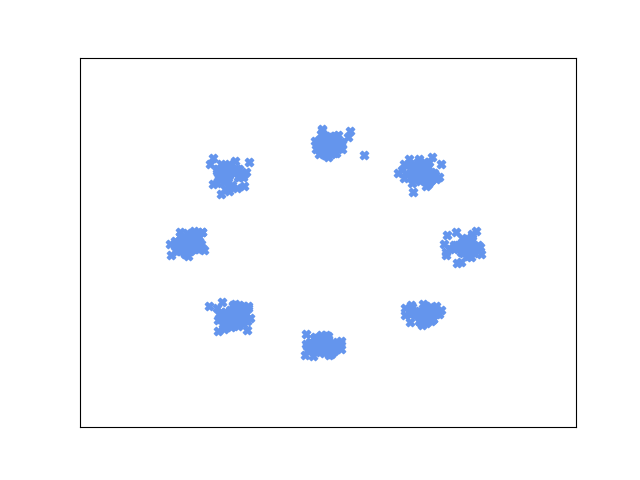}
    \end{subfigure}
    \begin{subfigure}{.19\textwidth}
    \includegraphics[width=\linewidth]{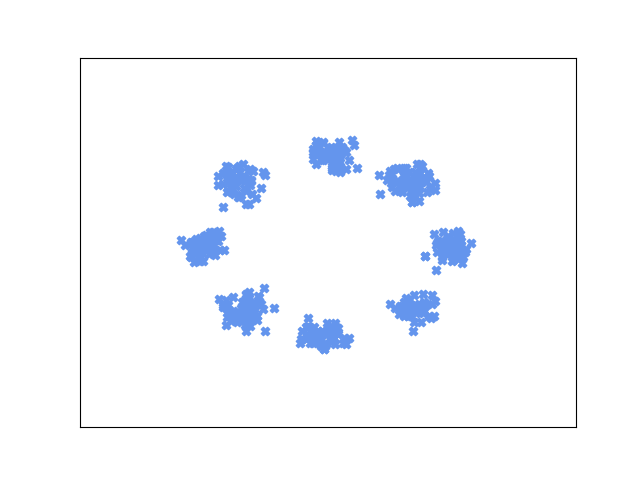}
    \end{subfigure}
    \begin{subfigure}{.19\textwidth}
    \includegraphics[width=\linewidth]{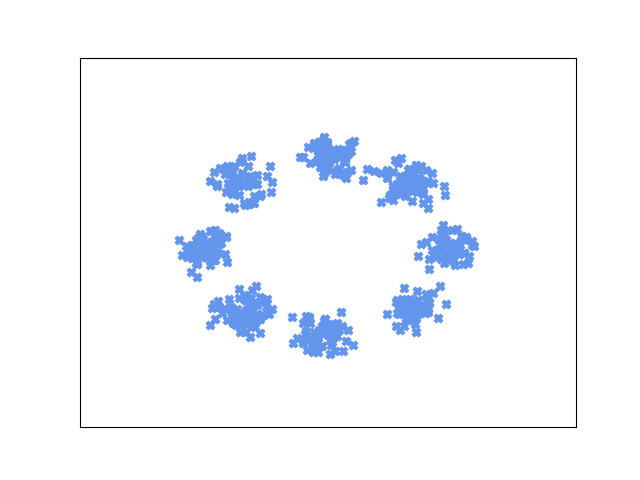}
    \end{subfigure}
    
    \begin{subfigure}{.19\textwidth}
    \includegraphics[width=\linewidth]{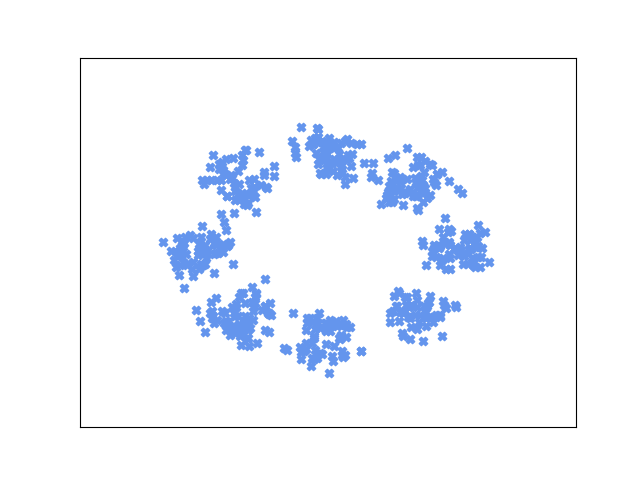}
    \end{subfigure}
    \begin{subfigure}{.19\textwidth}
    \includegraphics[width=\linewidth]{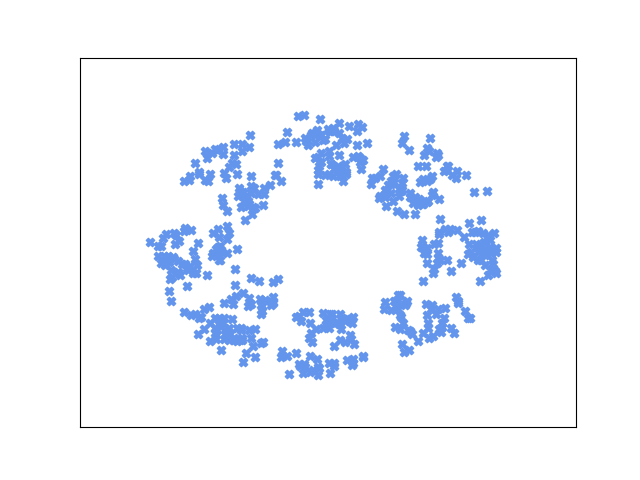}
    \end{subfigure}
    \begin{subfigure}{.19\textwidth}
    \includegraphics[width=\linewidth]{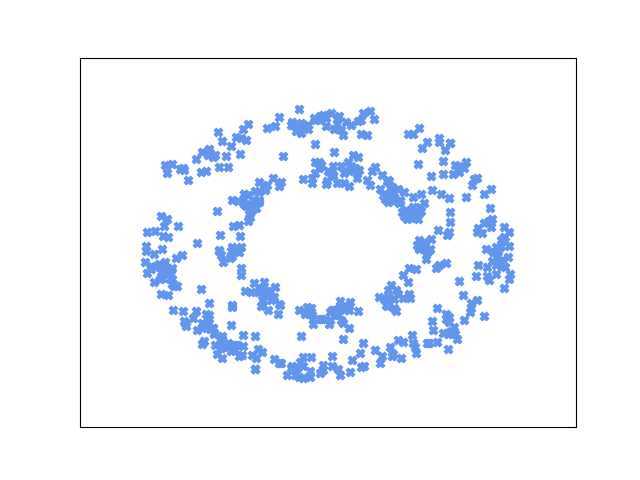}
    \end{subfigure}
    \begin{subfigure}{.19\textwidth}
    \includegraphics[width=\linewidth]{fig/toy/8gauss2circ/rsb/b_8.png}
    \end{subfigure}
\caption{Detailed qualitative results of RSB on data translation between 8-Gaussians and Circles of 2D Toy. The top two rows illustrate the forward process from Circles to 8-Gaussians. And other rows illustrate the backward process from 8-Gaussians to Circles.}
\label{fig:toy_8gauss2circ_rsb}
\end{center}
\end{figure*}

Since the SB-based formulation enables a stochastic process between any data spaces, the proposed RSB was tested on the data translation, \emph{i.e.} the case of conditional generation, between 8-Gaussians and Circles data space. Both RSB and DSB were trained with 8 timesteps, and 20K iterations for the bidirectional processes. And hyperparameters for RSB was set to $\alpha=0.5$ and $\beta=2.5$. And GAN was trained for 30K iterations.
See Figure \ref{fig:toy_8gauss2circ} for comparison. An interesting result is that when the conventional GAN was trained to generate the Circles data space from the 8-Gaussians data space, not the latent space, the training was not done properly.
And it can be confirmed that RSB obtained better translation performance than DSB. In addition, it can be visually confirmed that the trajectories of the trained stochastic process by RSB and DSB are different. See Figure \ref{fig:toy_8gauss2circ_dsb} and \ref{fig:toy_8gauss2circ_rsb}. 
When comparing the forward process from Circles to 8-Gaussians, the outer circle gathers in the form of 8-Gaussians and the central circle scatters toward the edge in DSB. In the case of RSB, the outer circle and the central circle are scattered and gathered to create 8 modes and they move to the desired place. Therefore, it can be concluded that applying regularization to the existing SB-based formulation changes the trajectories drawn by stochastic processes leading to fast training with smaller timesteps required.

These results from the 2D Toy show the possibility that the failure modes existing in both unconditional and conditional generation of GANs can be improved through stochastic-process-based generative models.

\subsubsection{Results with MNIST}

\begin{figure*}
\begin{center}
\centering
    \begin{subfigure}{.3\textwidth}
    \includegraphics[width=\linewidth]{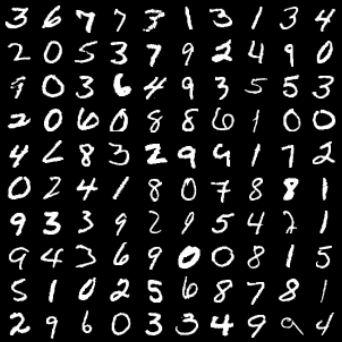}
    \caption{Ground Truth}
    \end{subfigure}
    \begin{subfigure}{.3\textwidth}
    \includegraphics[width=\linewidth]{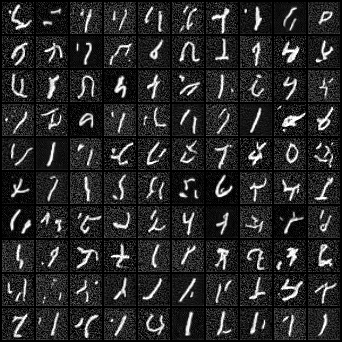}
    \caption{DSB}
    \end{subfigure}
    \begin{subfigure}{.3\textwidth}
    \includegraphics[width=\linewidth]{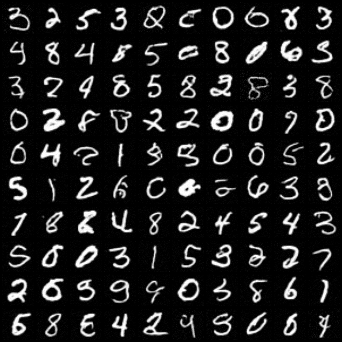}
    \caption{\textbf{RSB}}
    \end{subfigure}
\caption{Qualitative results of MNIST generation}
\label{fig:mnist}
\end{center}
\end{figure*}

With the MNIST dataset, RSB and DSB were compared for an unconditional generation task. Both RSB and DSB were trained with 16 timesteps, 8K iterations for the forward process, and 16K iterations for the backward process. And hyperparameters for RSB was set to $\alpha=0.5$ and $\beta=5$. The total iterations spent for training are relatively insufficient for training the existing SB models.
See Figure \ref{fig:mnist} for results. In the case of DSB, the training has not progressed significantly, but in the case of RSB, the training state has significantly progressed. It indicates that by adding regularization to the SB-based formulation, RSB reduces the number of iterations required for sufficient training compared to the existing SB models.

\subsubsection{Results with CelebA}
Recall that the advantage of SB-based formulation is that it can construct bidirectional stochastic processes between any two data spaces, eliminating the need for conditioning the generation starting from the latent space.
Therefore, the proposed RSB explored this possibility through the task of image-to-image translation and the single image super-resolution. 
The previous SB models \cite{chen2022likelihood, de2021diffusion} mainly demonstrated their performance on unconditional generation tasks, while the image size stayed in 32x32 size. But, since the RSB reduced the required number of timesteps and training time, it experimentally confirmed that the SB-based process could play a role in the relatively high-resolution data space of the CelebA dataset with 128x128 size.

\begin{figure*}
\begin{center}
\centering
    \begin{subfigure}{\textwidth}
    \centering
    \includegraphics[width=.2\linewidth]{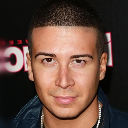}
    \includegraphics[width=.2\linewidth]{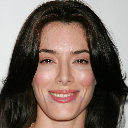}
    \includegraphics[width=.2\linewidth]{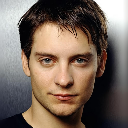}
    \includegraphics[width=.2\linewidth]{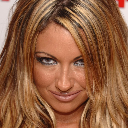}
    \caption{Source Image}
    \end{subfigure}
    \begin{subfigure}{\textwidth}
    \centering
    \includegraphics[width=.2\linewidth]{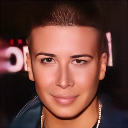}
    \includegraphics[width=.2\linewidth]{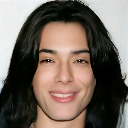}
    \includegraphics[width=.2\linewidth]{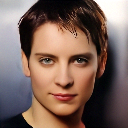}
    \includegraphics[width=.2\linewidth]{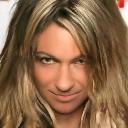}
    \caption{Translated Image}
    \end{subfigure}
\caption{Qualitative results on image-to-image translation task between male and female of CelebA. The top row illustrates the source images. And the bottom row illustrates the translated images.}
\label{fig:celebA_trans1}
\end{center}
\end{figure*}

The image-to-image translation task is to translate the image of the source data space to that of the target data space while maintaining the semantic information of the source image. There are various possible scenarios in image-to-image translation, and in this experiment, the translation between male and female faces was considered. 
The RSB trained the discrete-time stochastic processes between the data space of the male face and the female face. And the forward and backward processes were trained with $\alpha=0.5$, $\beta=10$, 4 timesteps, and 48K total iterations each.

\begin{figure*}
\begin{center}
\centering
    \begin{subfigure}{.19\textwidth}
    \includegraphics[width=\linewidth]{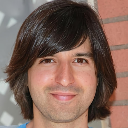}
    \end{subfigure}
    \begin{subfigure}{.19\textwidth}
    \includegraphics[width=\linewidth]{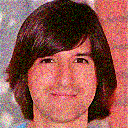}
    \end{subfigure}
    \begin{subfigure}{.19\textwidth}
    \includegraphics[width=\linewidth]{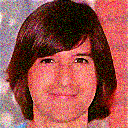}
    \end{subfigure}
    \begin{subfigure}{.19\textwidth}
    \includegraphics[width=\linewidth]{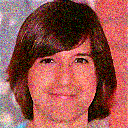}
    \end{subfigure}
    \begin{subfigure}{.19\textwidth}
    \includegraphics[width=\linewidth]{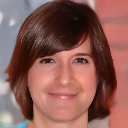}
    \end{subfigure}
\caption{Detailed translation process from male to female with 4 timesteps}
\label{fig:celebA_trans2}
\end{center}
\end{figure*}

The qualitative results from RSB are in Figure \ref{fig:celebA_trans1}. And Figure \ref{fig:celebA_trans2} shows the detailed translation process trained by the regularized stochastic process with 4 timesteps only.
The results show that the face of the source domain can be translated to the target domain while maintaining semantic information such as identity, facial expression, and pose. Note that masculinity or femininity was properly changed without a large change in hairstyles. It is seen as a result of SB-based formulation as an OT problem with implicit cycle-consistency \cite{de2019optimal, santambrogio2015optimal}. And it seems that implicit cycle-consistency was strongly applied.

\begin{figure*}
\begin{center}
\centering
    \begin{subfigure}{.19\textwidth}
    \includegraphics[width=\linewidth]{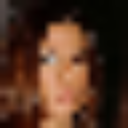}
    \end{subfigure}
    \begin{subfigure}{.19\textwidth}
    \includegraphics[width=\linewidth]{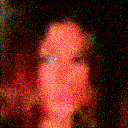}
    \end{subfigure}
    \begin{subfigure}{.19\textwidth}
    \includegraphics[width=\linewidth]{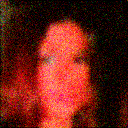}
    \end{subfigure}
    \begin{subfigure}{.19\textwidth}
    \includegraphics[width=\linewidth]{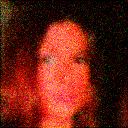}
    \end{subfigure}
    \begin{subfigure}{.19\textwidth}
    \includegraphics[width=\linewidth]{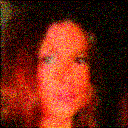}
    \end{subfigure}
    \begin{subfigure}{.19\textwidth}
    \includegraphics[width=\linewidth]{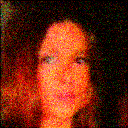}
    \end{subfigure}
    \begin{subfigure}{.19\textwidth}
    \includegraphics[width=\linewidth]{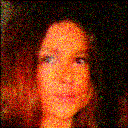}
    \end{subfigure}
    \begin{subfigure}{.19\textwidth}
    \includegraphics[width=\linewidth]{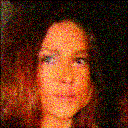}
    \end{subfigure}
    \begin{subfigure}{.19\textwidth}
    \includegraphics[width=\linewidth]{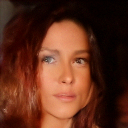}
    \end{subfigure}
\caption{Detailed super-resolution process from LR($\times8$) to SR}
\label{fig:celebA_SR}
\end{center}
\end{figure*}

Similarly, the RSB was trained to handle the single image super-resolution task. In this task, RSB constructed the stochastic processes between the data space of low-resolution images and super-resolution images. And it was trained with $\alpha=0.5$, $\beta=10$, 8 timesteps, and 48K total iterations for each forward and backward process.
The qualitative result and the detailed super-resolution process from RSB are in Figure \ref{fig:celebA_SR}. The desired super-resolution (SR) space was set to be 128x128 size and the low-resolution (LR) input was downsampled with $\times 8$ scale. The result shows that the desired SR output was attained while maintaining the semantic information of LR input properly.

\section{Conclusion}
\label{sec:conclusion}
This study tried to utilize bidirectional stochastic processes based on the Schrödinger bridge (SB) problem for deep generative modeling. The existing SB-based generative models have been proposed to improve the slow sampling speed of diffusion models and showed their potential. However, compared to generative models such as GANs, a large number of timesteps and a long training time are still required.
This study aimed to reduce the number of timesteps and training time required by the existing SB models. In the existing SB-based framework, the bidirectional stochastic processes become unstable with a small number of discretization timesteps because they are not consistent with each other. Therefore, this work proposed regularization terms to maintain the consistency between the bidirectional stochastic processes.
By applying this regularized SB-based process to both conditional and unconditional generation tasks, it was possible to properly train the desired stochastic processes between two arbitrary distributions even with smaller timesteps.

\bibliographystyle{unsrt}  
\bibliography{references}

\end{document}